%% file: ijcai22.tex
\documentclass{article}
\usepackage{ijcai22}

\usepackage{times}

\usepackage{soul}
\usepackage{url}
\usepackage[hidelinks]{hyperref}
\usepackage[utf8]{inputenc}
\usepackage[small]{caption}
\usepackage{booktabs}
\usepackage[normalem]{ulem}
\usepackage{natbib} 
\usepackage{nicefrac}       
\usepackage{xcolor}
\usepackage{colortbl} 
\usepackage{hhline}
\usepackage{multirow}
\usepackage{subfig}
\usepackage[american]{babel}
\usepackage{soul} 

\usepackage{tikz} 
\usepackage{amsfonts}
\usepackage{bm}
\usepackage{amsthm}
\usepackage{fancyhdr,graphicx,amsmath,amssymb}
\usepackage[ruled]{algorithm}
\usepackage{algcompatible}
\DeclareMathOperator*{\argmin}{arg\,min}
\DeclareMathOperator*{\argmax}{arg\,max}

\include{pythonlisting}

\newcommand{\LL}[1]{{\color{red}[LL: #1]}}

\newcommand{\EB}[1]{{\color{pink}[EB: #1]}}

\newtheorem{problem}{Problem}
\newtheorem{definition}{Definition}
\newtheorem{lemma}{Lemma}
\newtheorem{proposition}{Proposition}
\newtheorem{theorem}{Theorem}
\newtheorem{corollary}{Corollary}

\title{Individual Fairness Guarantees for Neural Networks}

\author{
Elias Benussi$^1$\footnote{Corresponding Author}\and
Andrea Patane$^1$\and
Matthew Wicker$^1$\and
Luca Laurenti$^2$\And
Marta Kwiatkowska$^1$\\
\affiliations
$^1$University of Oxford \\
$^2$TU Delft\\
\emails
$^1$\{elias.benussi, andrea.patane, matthew.wicker,  marta.kwiatkowska\}@cs.ox.ac.uk, $^2$l.laurenti@tudelft.nl
}

\begin{document}

\maketitle

\begin{abstract}
We consider the problem of certifying the individual fairness (IF) of feed-forward neural networks (NNs). 
In particular, we work with the $\epsilon$-$\delta$-IF formulation, which, given a NN and a similarity metric learnt from data, requires that the output difference between any pair of $\epsilon$-similar individuals is bounded by a maximum decision tolerance $\delta \geq 0$. 
Working with a range of metrics, including the Mahalanobis distance, we propose a method to overapproximate the resulting optimisation problem using piecewise-linear functions to lower and upper bound the NN's non-linearities globally over the input space.
We encode this computation as the solution of a Mixed-Integer Linear Programming problem and demonstrate that it can be used to compute IF guarantees on four datasets widely used for fairness benchmarking.
We show how this formulation can be used to encourage models' fairness at training time by modifying the NN loss, and empirically confirm our approach yields NNs that are orders of magnitude fairer than state-of-the-art methods.




\end{abstract}

\input{sections/1_intro}

\input{sections/2_problem_statement_rev}

\input{sections/3_verification}

\input{sections/4_training}

\input{sections/5_experiments}

\input{sections/6_conclusion}


\bibliography{ijcai22}

\clearpage
\newpage
\input{sections/7_appendix}

\end{document}

%% file: sections/1_intro.tex
\section{Introduction}\label{sec:intro}
Reservations have been raised about the application of neural networks (NN) in contexts where \textit{fairness} is of concern \citep{Barocas2018BigImpact}.
Because of inherent biases present in real-world data, if unchecked, these models have been found to discriminate against individuals on the basis of sensitive features, such as race or sex \citep{Bolukbasi2016ManEmbeddings, Angwin2016MachineBlacks}. 
Recently, the topic has come under the spotlight, with technologies being increasingly challenged for bias \citep{hardesty_2018, kirk2021OutOfTheBox, the_guardian_2020}, 
%
leading to the introduction of a range of definitions and techniques for capturing the multifaceted properties of fairness. 

Fairness approaches are broadly categorised  into: \textit{group fairness} \citep{Hardt2016EqualityLearning}, which inspects the model over data demographics; and \textit{individual fairness} (IF) \citep{Dwork2012FairnessAwareness}, which considers the behaviour over each individual.
Despite its wider adoption, group fairness is only concerned with statistical properties of the model so that a situation may arise where predictions of a group-fair model can be perceived as unfair by a particular individual.
In contrast, IF is a worst-case measure with guarantees over \textit{every} possible individual in the input space.
However, while techniques exist for group fairness of NNs \citep{albarghouthi2017fairsquare, Bastani2018ProbabilisticConcentration},  research on IF has thus far been limited to designing training procedures that favour fairness \citep{yurochkin2020training, yeom2020individual, McNamara2017ProvablyRepresentations} and verification over specific individuals \citep{ruoss2020learning}. To the best of our knowledge, there is currently no work targeted at global certification of IF for NNs.


We develop an anytime algorithm with provable bounds for the certification of IF on NNs. 
We build on the $\epsilon$-$\delta$-IF formalisation employed by \cite{john2020verifying}.
That is, given $\epsilon, \delta \geq 0$ and a distance metric $d_\text{fair}$ that captures the similarity between individuals, we ask that, for \textit{every} pair of points $x'$ and $x''$ in the input space with $d_\text{fair}(x',x'') \leq \epsilon $, the NN's output does not differ by more than $\delta$. 
Although related to it, IF certification on NNs poses a different problem than  adversarial robustness \citep{Tjeng2019EvaluatingProgramming},  as both $x'$ and $x''$ are here problem variables, spanning the whole space{. Hence, local approximation techniques developed in the adversarial literature cannot be employed in the context of IF.}

Nevertheless, we show how this global, non-linear requirement can be encoded in Mixed-Integer Linear Programming (MILP) form, by deriving a set of global upper and lower piecewise-linear (PWL) bounds over each activation function in the NN over the whole input space, and performing linear encoding of the (generally non-linear) similarity metric  $d_\text{fair}(x',x'')$. 
The formulation of our optimisation as a MILP allows us to compute an \textit{anytime}, worst-case bound on IF, which can thus be computed using standard solvers from the global optimisation literature \citep{dantzig2016linear}. 
%
Furthermore, we demonstrate how our approach can be embedded into the NN training so as to optimise for individual fairness at training time.
We do this by performing gradient descent on a weighted loss that also accounts for the maximum $\delta$-variation in $d_{\text{fair}}$-neighborhoods for each training point, similarly to what is done in adversarial learning  \citep{goodfellow2014explaining, Gowal2018OnModels,wicker2021bayesian}.

We apply our method on four benchmarks widely employed in the fairness literature, namely, the Adult, German, Credit and Crime datasets \citep{UCIDatasets}, and  an array of similarity metrics learnt from data that include $\ell_\infty$, Mahalanobis, and NN embeddings. 
We empirically demonstrate how our method is able to provide the first, non-trivial IF certificates for NNs commonly employed for tasks from the IF literature, and even larger NNs comprising up to thousands of neurons.  
Furthermore, we find that 
our MILP-based fair training approach consistently outperforms, in terms of IF guarantees, NNs trained with a competitive state-of-the-art 
technique by orders of magnitude, albeit at an increased computational cost.

The paper makes the following main contributions:\footnote{Proofs and additional details can be found in Appendix of an extended version of the paper available at \url{http://www.fun2model.org/bibitem.php?key=BPW+22}.}
\begin{itemize}
    \item We design a MILP-based, anytime verification approach for the certification of IF as a global property on NNs.
    \item We demonstrate how our technique can be used to modify the loss function of a NN to take into account certification of IF at training time.
    \item On four datasets, and an array of metrics, we show how our techniques obtain non-trivial IF certificates and train NNs that are significantly fairer than state-of-the-art. 
\end{itemize}
\paragraph{Related Work}
A number of works have considered IF by employing techniques from adversarial robustness.
\cite{yeom2020individual}  rely on randomized smoothing to find the highest stable per-feature difference in a model. Their method, however, provides only (weak) guarantees on model statistics.
\cite{yurochkin2020training} present  a method for IF training that builds on projected gradient descent and optimal transport. While the method is found to decrease model bias to state-of-the-art results, no formal guarantees are obtained.
\cite{ruoss2020learning} adapted the MILP formulation for adversarial robustness to handle fair metric embeddings.
However, rather than tackling the IF problem globally as introduced by \cite{Dwork2012FairnessAwareness}, the method only works iteratively on a finite set of data, hence leaving open the possibility of unfairness in the model. In contrast, the MILP encoding we obtain through PWL bounding of activations and similarity metrics allows us to provide guarantees over \textit{any} possible pair of individuals.
\cite{Urban2020PerfectlyNetworks} employ static analysis to certify causal fairness. While this method yields global guarantees, it cannot be straightforwardly employed for IF, and  it is not \textit{anytime}, making exhaustive analysis impractical.
\cite{john2020verifying} present a method for the computation of IF, though limited to linear and kernel models. 
MILP and linear relaxation have been employed to certify NNs in local adversarial settings
\citep{ehlers2017formal,Tjeng2019EvaluatingProgramming,wicker2020probabilistic}.
However, local approximations cannot be employed for the global IF problem. 
%
While \cite{katz2017reluplex,leino2021globally} consider global robustness, their methods are restricted to $\ell_p$ metrics. Furthermore, they require the knowledge of a Lipschitz constant or are limited to ReLU.

%% file: sections/2_problem_statement_rev.tex
\section{Individual Fairness}\label{sec:IndFairness}


We focus on regression and binary classification with NNs with real-valued inputs and one-hot encoded categorical features.\footnote{Multi-class can be tackled with component-wise analyses.}
Such frameworks are often used in automated decision-making, e.g.\ for loan applications \citep{Hardt2016EqualityLearning}. 
%
%
Formally, 
given a compact input set $X\subseteq \mathbb{R}^n$ and an output set $Y\subseteq \mathbb{R}$, we consider an $L$ layer fully-connected NN $f^{w} : X\rightarrow Y$,  parameterised by a vector of weights  $w \in \mathbb{R}^{n_w}$ trained on
$\mathcal{D}=\{(x_i,y_i), i\in \{1,...,n_{d} \}\}$. 
 For an input $x \in X$, $i=1,\ldots,L$ and $j=1,\ldots,n_i$,  the NN is defined as: 
\begin{align}
    &\phi^{(i)}_j = \sum_{k =1}^{n_{i-1}} W^{(i)}_{jk} \zeta^{(i-1)}_k  + b^{(i)}_{j},  
    &\zeta^{(i)}_j = \sigma^{(i)}\left(\phi^{(i)}_j\right) \label{eq:nn}
\end{align}
where $\zeta^{(0)}_j = x_j$.
Here, $n_i$ is the number of units in the $i$th layer, $W^{(i)}_{jk}$ and  $b^{(i)}_{j}$ are its weights and biases, $\sigma^{(i)}$ is the activation function, $\phi^{(i)}$ is the pre-activation and $\zeta^{(i)}$ the activation. 
The NN output is the result of these computations, $f^w(x) := \zeta^{(L)}.$ 
In regression, $f^w(x)$ is the prediction, while for classification it represents the class probability. In this paper we focus on fully-connected NNs as widely employed in the IF literature \citep{yurochkin2020training, Urban2020PerfectlyNetworks,ruoss2020learning}. However, we should stress that our framework,  being based on MILP, can be easily extended to convolutional, max-pool and batch-norm layers or res-nets by using embedding techniques from the adversarial robustness literature (see e.g. \citep{boopathy2019cnn}.


\paragraph{Individual Fairness} 
Given a NN  $f^w$, IF \citep{Dwork2012FairnessAwareness}  enforces the property that 
similar individuals are similarly treated. 
Similarity is defined according to a task-dependent  pseudometric, $d_\text{fair}: X \times X  \mapsto \mathbb{R}_{\geq 0}$, provided by a domain expert (e.g., a Mahalanobis distance correlating each feature to the sensitive one),  whereas similarity of treatment is expressed via the absolute difference on the NN output $f^w(x)$. 
%
We adopt the $\epsilon$-$\delta$-IF formulation of \cite{john2020verifying} for the formalisation of input-output IF similarity.

%
\begin{definition}[$\epsilon$-$\delta$-IF \citep{john2020verifying}]\label{def:IndFairness}
Consider $\epsilon \geq 0$ and $\delta \geq 0$. We say that $f^w$ is $\epsilon$-$\delta$-individually fair w.r.t.\ $d_{\text{fair}}$ iff
\begin{align*}
 \forall x',x'' \; \text{s.t.} \;    d_{\text{fair}}(x',x'') \leq \epsilon \implies |f^w(x') - f^w(x'')| \leq  \delta.
\end{align*}
\end{definition}
%
Here, $\epsilon$ measures similarity between individuals and $\delta$ is the difference in outcomes (class probability for classification).
We emphasise that individual fairness is a \textit{global} notion, as the condition in Definition \ref{def:IndFairness} must hold for all pairs of points in $X$.
We remark that the $\epsilon$-$\delta$-IF formulation of \cite{john2020verifying} (which is more general than IF formulation typically used in the literature \citep{yurochkin2020training, ruoss2020learning}) is a slight variation on the Lipschitz property introduced by \cite{Dwork2012FairnessAwareness}.
While introducing greater flexibility thanks to its parametric form, it makes an IF parametric analysis necessary at test time.
In Section \ref{sec:ExperimentalResults} we analyse how $\epsilon$-$\delta$-IF of NNs is affected by variations of $\epsilon$ and $\delta$.
%
%
A crucial component of IF is the similarity  $d_\text{fair}$. 
The intuition is that sensitive features, or their sensitive combination, should not influence the NN output. 
While a number of metrics has been discussed in the literature \citep{ilvento19metric}, we focus on the following representative set of metrics 
which can be automatically learnt from data \citep{john2020verifying,ruoss2020learning,mukherjee20simple,yurochkin2020training}. Details on metric learning is given in Appendix~\ref{sec:metric_learning}.\\
\textbf{Weighted $\ell_p$:}
In this case $d_{\text{fair}}(x',x'')$ is defined as a weighted version of an $\ell_p$ metric, i.e.\
$d_{\text{fair}}(x',x'') = \sqrt[p]{\sum_{i=1}^n \theta_i |x'_i - x''_i|^p}  $.
Intuitively, we set the weights $\theta_i$ related to  sensitive features to zero, so that two individuals are considered similar if they only differ with respect to those.
The weights $\theta_i$ for the remaining features can be tuned according to their degree of correlation to the sensitive features.\\
\textbf{Mahalanobis:}
In this case we have $d_{\text{fair}}(x',x'')  = \sqrt{(x'-x'')^T S (x'-x'')}$, for a given positive semi-definite (SPD) matrix $S$.
The Mahalanobis distance generalises the $\ell_2$ metric by taking into account the intra-correlation of features to capture latent dependencies w.r.t.\ the sensitive features.\\ 
\textbf{Feature Embedding:} 
The metric is computed on an embedding, so that $d_{\text{fair}}(x',x'') = \hat{d}(\varphi(x'),\varphi(x''))$, where $\hat{d}$ is either the Mahalanobis or the weighted  $\ell_p$ metric, and  $\varphi$ is a feature embedding map. These allow for greater modelling flexibility, at the cost of reduced interpretability.



\subsection{Problem Formulation}
We aim at certifying $\epsilon$-$\delta$-IF for NNs.
To this end we
formalise two problems: computing certificates and training for IF.
\begin{problem}[Fairness Certification]\label{prob:certification}
Given a trained NN $f^w$, a similarity $d_{\text{fair}}$ and a distance threshold $\epsilon \geq 0$, compute 
\begin{align*}
    \delta_{\max}=\max_{\substack{ x',x'' \in X \\  d_{\text{fair}}(x',x'') \leq \epsilon}} |f^w(x') - f^w(x'')|.
\end{align*}
\end{problem}
Problem \ref{prob:certification} provides a formulation in terms of optimisation, 
seeking to compute the maximum  output change $\delta_{\max}$ for any pair of input points  whose $d_{\text{fair}}$ distance is no more than $\epsilon$. One can then compare $\delta_{\max}$ with any threshold $\delta$: if $\delta_{\max} \le \delta$ holds then the model $f^w$ has been certified to be $\epsilon$-$\delta$-IF. 

While Problem~\ref{prob:certification} is concerned with an already trained NN, the methods we develop can also be employed to encourage IF at training time.   
Similarly to the approaches for adversarial learning \citep{goodfellow2014explaining}, we modify the training loss $L(f^w(x),y)$ to balance between the  model fit and IF.  
\begin{problem}[Fairness Training]\label{prob:training}
Consider an NN $f^w$, a training set $\mathcal{D}$, a similarity metric $d_{\text{fair}}$ and a distance threshold $\epsilon \geq 0$. Let $\lambda \in [0,1]$ be a constant. Define the IF-fair loss as 
\begin{align*}
    L_{\text{fair}}&(f^w(x_i),y_i,f^w(x^*_i),\lambda)  = \\ &\lambda L(f^w(x_i),y_i) + (1-\lambda) | f^w(x_i) - f^w(x^*_i) |,
\end{align*}
where $x^*_i=\argmax_{x \in X \, s.t.\, d_{\text{fair}}(x_i,x) \leq \epsilon} |f^w(x_i) - f^w(x)|$. 
The $\epsilon$-IF training problem is defined as finding $w^{\text{fair}}$ s.t.:
\begin{align*}
w^{\text{fair}}   = \argmin_{w}\sum_{i = 1}^{n_d} L_{\text{fair}}(f^w(x_i),y_i). 
\end{align*}
\end{problem}
In Problem~\ref{prob:training} we seek to train a NN that not only is accurate, but whose predictions are also fair according to Definition~\ref{def:IndFairness}. Parameter $\lambda$ balances between accuracy and IF. In particular, for $\lambda=1$ we recover the standard training that does not account for IF, while for $\lambda=0$ we only consider IF.

%% file: sections/3_verification.tex
\section{A MILP Approach For Individual Fairness}\label{sec:MILP}
Certification of individual fairness on a NN thus requires us to solve the following global, non-convex optimisation problem:
\begin{align}
    \max_{x',x'' \in X}&\quad \vert \delta \vert \nonumber\\
    \text{subject to} & \quad \delta = f^w(x') - f^w(x'') \label{eq:obj_function}\\
    & \quad  d_{\text{fair}}(x',x'') \leq \epsilon. \label{eq:con_function}
\end{align}
We develop a Mixed-Integer Linear Programming (MILP) over-approximation (i.e., providing a sound bound) to this problem. 
We notice that there are two sources of non-linearity here, one induced by the NN (Equation \eqref{eq:obj_function}), which we refer to as  the \textit{model constraint}, and the other by the fairness metric (Equation \eqref{eq:con_function}), which we call \textit{fairness constraint}. 
In the following, we show how these can be modularly bounded by piecewise-linear functions. In Section \ref{sec:overall_formulation} we bring the results together to derive a MILP formulation for $\epsilon$-$\delta$-IF. 
\begin{figure*}[h]
    \centering
    \includegraphics[width=0.32\textwidth]{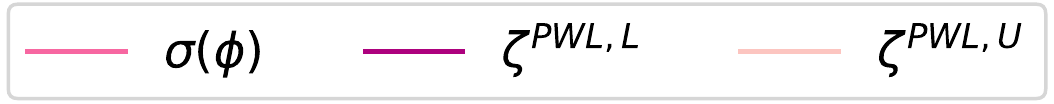} \\
    \includegraphics[width=0.2\textwidth]{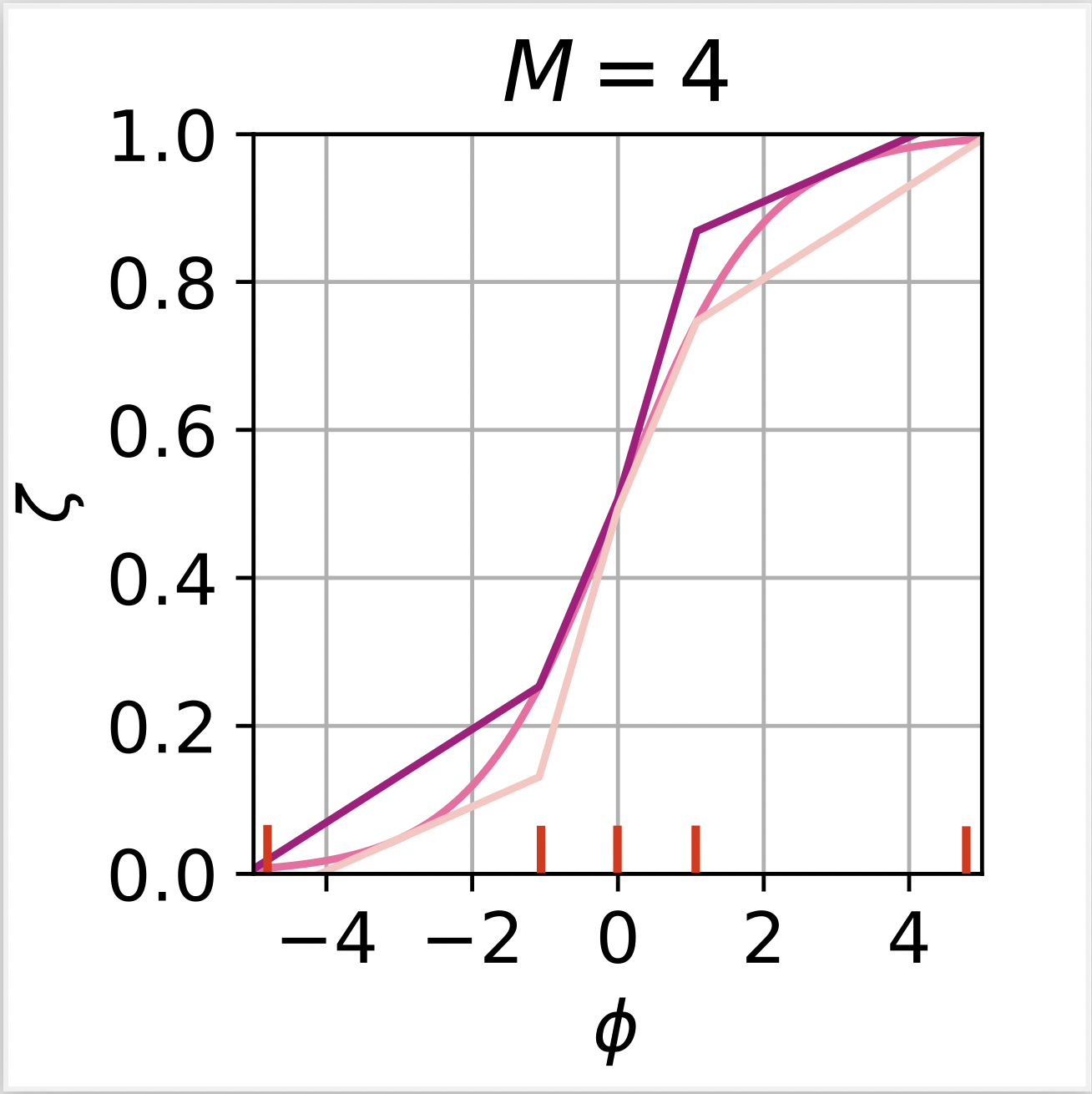}
    \includegraphics[width=0.2\textwidth]{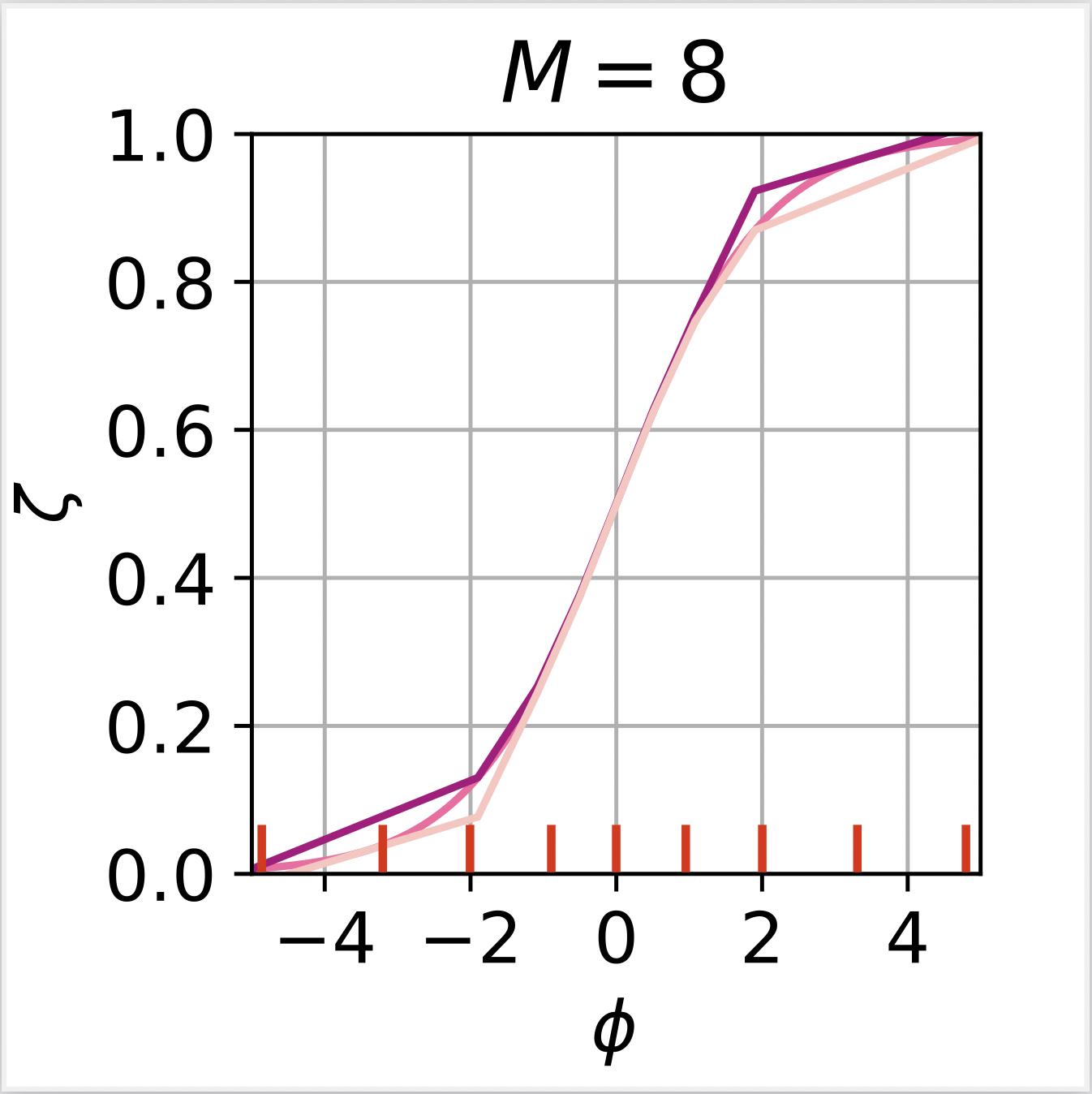}
    \includegraphics[width=0.2\textwidth]{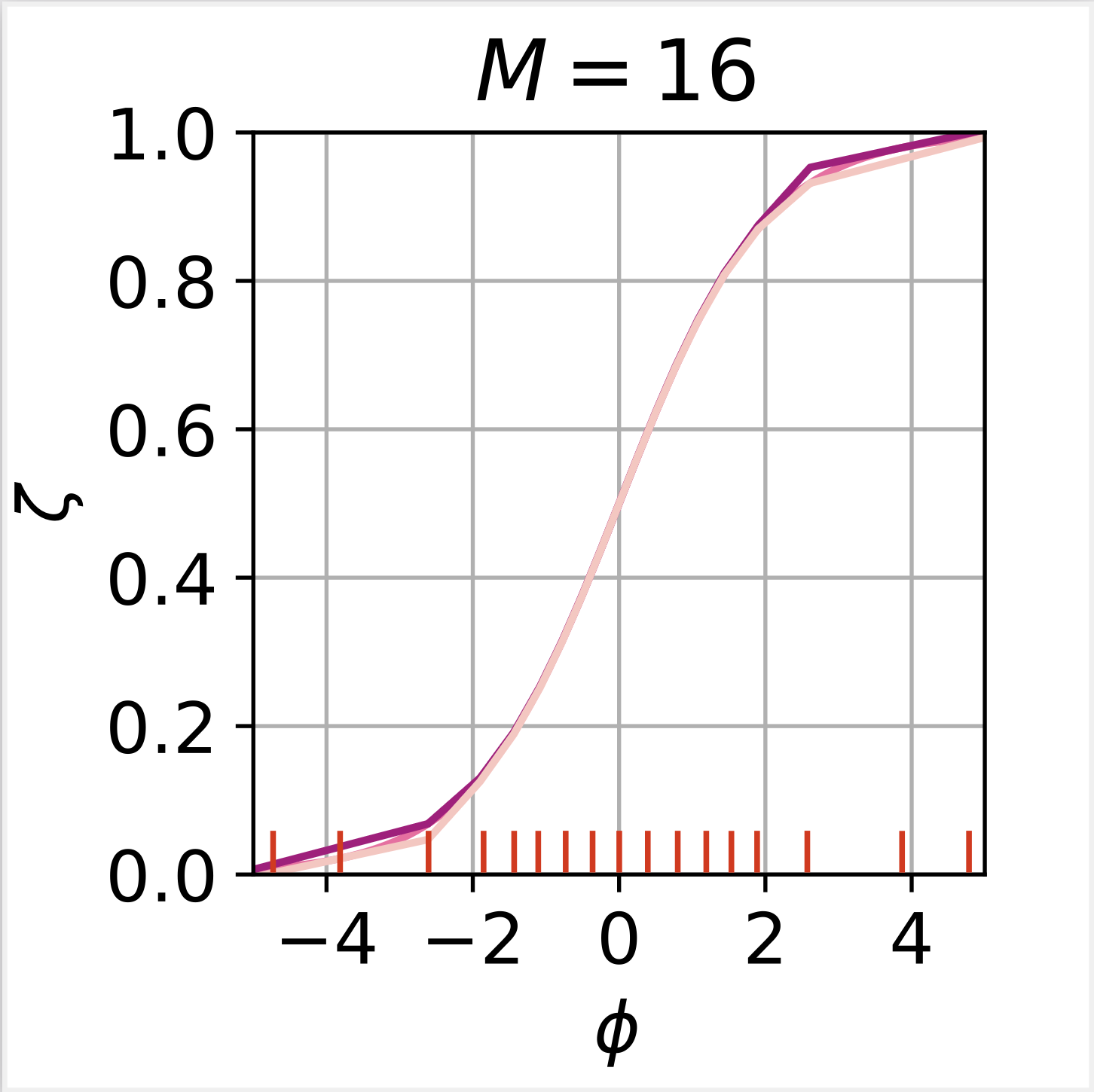}
    \includegraphics[width=0.2\textwidth]{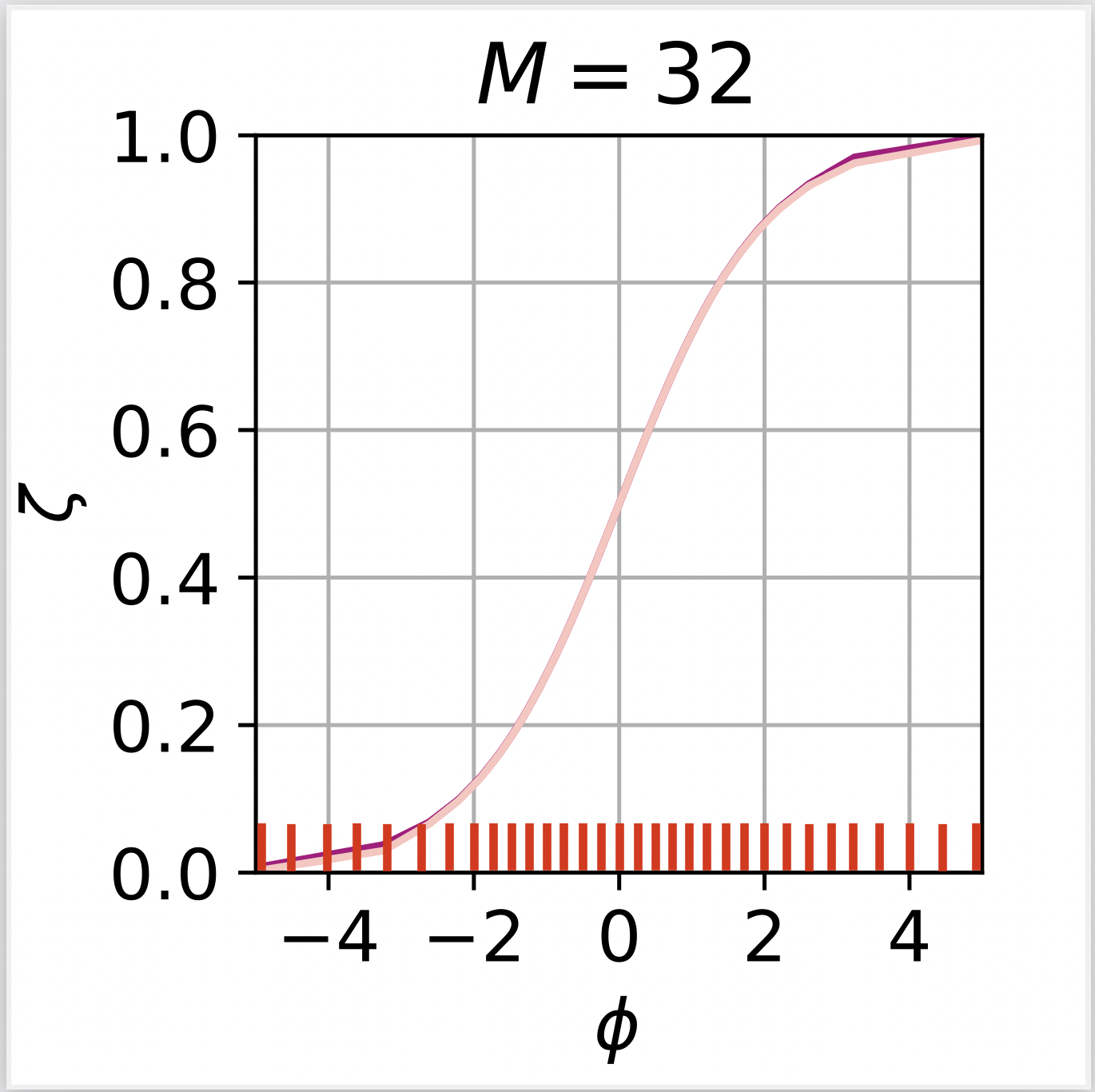}
    \caption{Upper and lower PWL functions to sigmoid for an increasing number of partition points $M$ (marked with red ticks).}
    \label{fig:sigmoid_discretisation}
\end{figure*}
\subsection{Model Constraint}\label{sec:model_constraint}
%
%
We develop a scheme based on \textit{piecewise-linear} (PWL) upper and lower bounding for over-approximating all commonly used non-linear activation functions. 
%
An illustration of the PWL bound is given in Figure \ref{fig:sigmoid_discretisation}.
Let $\phi^{(i)L}_j$ and $\phi^{(i)U}_j \in \mathbb{R}$ be lower and upper bounds on the pre-activation $\phi^{(i)}_j$.\footnote{Computed  by bound propagation over $X$ \citep{Gowal2018OnModels}.}
We proceed by building a discretisation grid over the $\phi^{(i)}_j$ values on $M$ grid points: 
$\bm{\phi}_{\text{grid}} = [ \phi^{(i)}_{j,0}, \ldots,  \phi^{(i)}_{j,M} ]  $, with $\phi^{(i)}_{j,0}:=\phi^{(i)L}_j$ and $\phi^{(i)}_{j,M}:=\phi^{(i)U}_j$,  such that, in each partition interval $[\phi^{(i)}_{j,l} ,\phi^{(i)}_{j,l+1}]$, we have that $\sigma^{(i)}$ is either convex or concave. 
We then compute linear lower and upper bound functions for $\sigma^{(i)}$ in each $[\phi^{(i)}_{j,l} ,\phi^{(i)}_{j,l+1}]$ as follows. 
%
%
If $\sigma^{(i)}$ is convex (resp.\ concave) in $[\phi^{(i)}_{j,l} ,\phi^{(i)}_{j,l+1}]$, then an upper (resp.\ lower) linear bound is given by the segment connecting the two extremum points of the interval, and a lower (resp.\ upper) linear bound is given by the tangent through the mid-point of the interval.
We then compute the values of each linear bound in each of its grid points, and select the minimum of the lower bounds and the maximum of the upper bound values, which we store in two vectors $\bm{\zeta}^{\text{PWL},(i),U}_{j} = [ \zeta^{\text{PWL},(i),U}_{j,0}, \ldots,  \zeta^{\text{PWL},(i),U}_{j,M} ]  $ and $\bm{\zeta}^{\text{PWL},(i),L}_{j} = [ \zeta^{\text{PWL},(i),L}_{j,0}, \ldots,  \zeta^{\text{PWL},(i),L}_{j,M} ]$.
The following lemma is a consequence of this construction.
\begin{lemma}
Let $\phi \in [\phi^{(i)L}_j,\phi^{(i)U}_j]$. Denote with $l$ the index associated to the partition of $\bm{\phi}_{\text{grid}}$ in which $\phi$ falls and consider $\eta \in [0,1]$ such that $\phi =\eta \phi^{(i)L}_{j,l-1} + (1-\eta) \phi^{(i)L}_{j,l} $. Then:
\begin{align*}
\sigma^{(i)}(\phi) \geq \eta  \zeta^{\text{PWL},(i),L}_{j,l-1} + (1-\eta) \zeta^{\text{PWL},(i),L}_{j,l}, \\ \sigma^{(i)}(\phi)  \leq \eta  \zeta^{\text{PWL},(i),U}_{j,l-1} + (1-\eta) \zeta^{\text{PWL},(i),U}_{j,l},
\end{align*} 
that is, $\bm{\zeta}^{\text{PWL},(i),L}_{j}$ and $\bm{\zeta}^{\text{PWL},(i),U}_{j}$ define continuous PWL lower and upper bounds for $\phi$ in $ [\phi^{(i)L}_j,\phi^{(i)U}_j]$.
\end{lemma}\label{lemma:pwl_boudning}
Lemma \ref{lemma:pwl_boudning} guarantees that we can bound the non-linear activation functions using PWL functions.
Crucially, PWL functions can then be encoded into the MILP constraints. 
\begin{proposition}\label{proposition:milp_pwl}
Let $y^{(i)}_{j,l}$ for $l=1,\ldots,M$, be binary variables, and $\eta^{(i)}_{j,l} \in [0,1]$ be continuous ones.
Consider $\phi^{(i)}_j \in [\phi^{(i)L}_j,\phi^{(i)U}_j] $ then it follows that  $\zeta^{(i)}_j = \sigma^{(i)}\left(\phi^{(i)}_j\right)$ implies:
%
\begin{align*}
     &\sum_{l=1}^M y^{(i)}_{j,l} = 1, \; \sum_{l=1}^M \eta^{(i)}_{j,l} = 1, \phi^{(i)}_j = \sum_{l=1}^M \phi^{(i)L}_{j,l} \eta^{(i)}_{j,l}, \;
     y^{(i)}_{j,l}  \leq \\ & \eta^{(i)}_{j,l} +  \eta^{(i)}_{j,l+1}, \;
     \sum_{l=1}^M  \zeta^{\text{PWL},(i),L}_{j,l}\eta^{(i)}_{j,l} \leq \zeta^{(i)}_j \leq \sum_{l=1}^M  \zeta^{\text{PWL},(i),U}_{j,l}\eta^{(i)}_{j,l}. 
\end{align*}
\end{proposition}

\begin{figure*}[ht]
\centering
  \centering
  \includegraphics[width=0.8\textwidth]{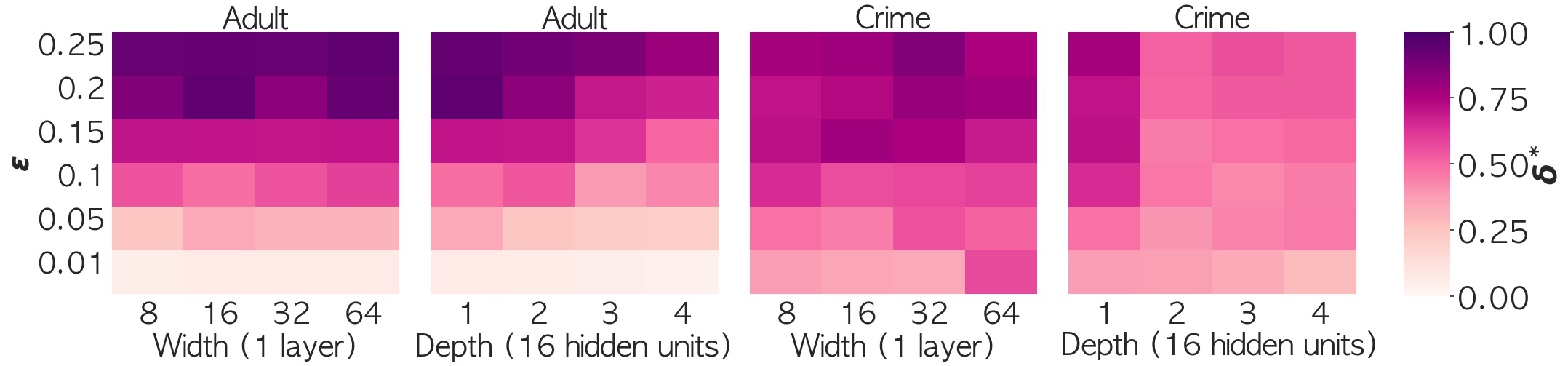}\label{fig:verification-mahalanobis} \\
  \includegraphics[width=0.8\textwidth]{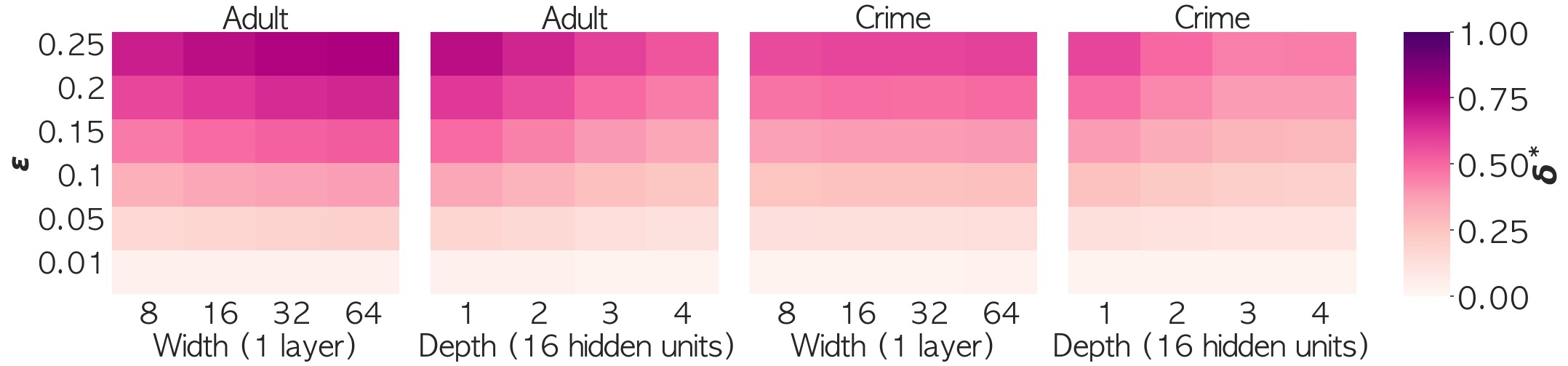}\label{fig:verification-weighted-lp}
  \caption{Certified bounds on IF ($\delta_*$) for different architecture parameters (widths and depths) and maximum similarity ($\epsilon$) for the Adult and the Crime datasets. \textbf{Top Row}: Mahalanobis metric used for $d_{\text{fair}}$. \textbf{Bottom Row}: Weighted $\ell_{\infty}$ metric used for $d_{\text{fair}}$.}
  \label{fig:verification-fig}
\end{figure*}

A proof  can be found in Appendix A.
Proposition \ref{proposition:milp_pwl} ensures that the global behaviour of each NN neuron can be over-approximated by $5$ linear constraints using $2M$ auxiliary variables. Employing Proposition \ref{proposition:milp_pwl} we can encode the model constraint of Equation \eqref{eq:obj_function} into the MILP form in a sound way.

The over-approximation error does not depend on the MILP formulation (which is exact), but on the PWL bounding, and is hence controllable through the selection of the number of grid points $M$, and becomes  exact in the limit. Notice that in the particular case of ReLU activation functions the over-approximation is exact for any $M>0.$

\begin{proposition}\label{proposition:M_convergence}
Assume $\sigma^{(i)}$ to be continuously differentiable everywhere in $[\phi^{(i)L}_j,\phi^{(i)U}_j]$, except possibly in a finite set.
Then PWL lower and upper bounding functions of Lemma \ref{lemma:pwl_boudning} converge uniformly to $\sigma^{(i)}$  as $M$ goes to infinity.

Furthermore, define $\Delta_M = (\phi^{(i)U}_j - \phi^{(i)L}_j) /M $, then  for finite values of $M$ the error on the lower (resp.\ upper) bounding in convex (resp.\ concave) regions of $\sigma^{(i)}$ for $\phi \in [\phi^{(i)}_{j,l}, \phi^{(i)}_{j,l+1} ]$   is given by:
\begin{align*}
    e_1(\phi) \leq \frac{\Delta_M}{2} \left( \sigma' (\phi^{(i)}_{j,l+1}) - \sigma'\left(\phi^{(i)}_{j,l+1} -  \frac{\Delta_M}{2}  \right) \right)
\end{align*}
and upper (resp.\ lower) in concave (resp.\ convex) regions:
\begin{align*}
     e_2(\phi) \leq \Delta_M \left( \frac{\sigma \left(\phi^{(i)}_{j,l} + \Delta_M \right) - \sigma(\phi^{(i)}_{j,l})}{ \Delta_M } + \sigma'(\phi^{(i)}_{j,l} )\right) .
\end{align*}
\end{proposition}

A proof of Proposition \ref{proposition:M_convergence} is given in Appendix~\ref{sec:milp_details}, alongside an experimental analysis of the convergence rate. 

We remark that the PWL bound can be used over all commonly employed activation functions $\sigma$. The only assumption made is that $\sigma$ has a finite number of inflection points  over any compact interval of $\mathbb{R}$. For convergence (Prop.\ 2) we require continuous differentiability almost everywhere, which is satisfied by commonly used activations.

\subsection{Fairness Constraint}\label{sec:fairness_constraint}
The encoding of the fairness constraint within the MILP formulation depends on the specific form of the metric $d_{\text{fair}}$.

\noindent\textbf{Weighted $\ell_p$ Metric}:
The weighted $\ell_p$ metric can be tackled by employing rectangular approximation regions. While this is straightforward for the $\ell_\infty$ metric, for the remaining cases interval abstraction can be used \citep{dantzig2016linear}.

\noindent\textbf{Mahalanobis Metric}:
We first compute an orthogonal decomposition of $S$ as in  $U^T S U  = \Lambda$, where $U$ is the eigenvector matrix of $S$ and $\Lambda$ is a diagonal matrix with $S$ eigenvalues as entries.
Consider the rotated variables  $z' = U^Tx'$ and $z'' = U^Tx''$, then we have that Equation \eqref{eq:con_function} can be re-written as $(z' - z'')^T \Lambda  (z' - z'') \leq \epsilon^2$.
By simple algebra we thus have that, for each $i$, $(z'_i - z''_i)^2 \leq \frac{\epsilon^2}{\Lambda_{ii}}$. 
By transforming back to the original variables, we obtain that Equation \eqref{eq:con_function} can be over-approximated by:
%
    $ - \frac{\epsilon}{\sqrt{\text{diag}(\Lambda)}} \leq U^Tx' - U^Tx'' \leq \frac{\epsilon}{\sqrt{\text{diag}(\Lambda)}}.$
%

\noindent\textbf{Feature Embedding Metric}
We tackle the case in which $\varphi$ used in the metric definition, i.e. $d_{\text{fair}}(x',x'') = \hat{d}(\varphi(x'),\varphi(x''))$, is a NN embedding.
This is straightforward as $\varphi$ can be encoded into MILP as for the model constraint.

\subsection{Overall Formulation}\label{sec:overall_formulation}
We now formulate the MILP encoding for the over-approximation $\delta_* \geq \delta_{max}$ of $\epsilon$-$\delta$-IF.
For Equation \eqref{eq:obj_function}, we proceed by deriving a set of approximating constraints for the variables $x'$ and $x''$ by using the techniques described in Section \ref{sec:model_constraint}.
We denote the corresponding variables as $\phi'^{(i)}_j$, $\zeta'^{(i)}_j$ and  $\phi''^{(i)}_j$, $\zeta''^{(i)}_j$, respectively. 
The NN final output on $x'$ and on $x''$ will then respectively be $\zeta'^{(L)}$ and $\zeta''^{(L)}$, so that $\delta = \zeta'^{(L)} - \zeta''^{(L)}$.
Finally, we over-approximate Equation \eqref{eq:con_function} as described in Section \ref{sec:fairness_constraint}. 
In the case of Mahalanobis distance, we thus obtain: 
{
\begin{align} \label{eq:milp}
    \max_{x',x'' \in X} & \quad | \delta | \\
    \text{subject} & \text{to}
     \; = \zeta'^{(L)} - \zeta''^{(L)} \nonumber \\
    & \text{for}\;  i=1,\ldots,L,  \;\; j=1,\ldots,n_i,\; \dag \in \{','' \}: \nonumber \displaybreak[0]  \\ 
    & \;  \sum_{l=1}^M y^{\dag(i)}_{j,l} = 1,  \quad \sum_{l=1}^M \eta^{\dag(i)}_{j,l} = 1, \nonumber \displaybreak[0] \quad  y^{(i)}_{j,l}  \leq \eta^{(i)}_{j,l}  + \eta^{(i)}_{j,l+1}\\
    & \;  \phi^{\dag (i)}_j = \sum_{ k=1 }^{n_{i-1}}W^{(i)}_{jk}x^\dag_k + b^{(i)}_j, \; \phi^{\dag(i)}_j = \sum_{l=1}^M \phi^{(i)L}_{j,l} \eta^{\dag(i)}_{j,l} \nonumber \displaybreak[0]\\
    & \;  \sum_{l=1}^M  \zeta^{\text{PWL},(i),L}_{j,l}\eta^{\dag(i)}_{j,l} \leq \zeta^{\dag(i)}_j \leq \sum_{l=1}^M  \zeta^{\text{PWL},(i),U}_{j,l}\eta^{\dag(i)}_{j,l}  \nonumber  \displaybreak[0]\\
    & - \frac{\epsilon^2}{\sqrt{\text{diag}(\Lambda)}} \leq Ux' - Ux'' \leq \frac{\epsilon^2}{\sqrt{\text{diag}(\Lambda)}}. \nonumber 
\end{align}
}%

Though similar, the above MILP is significantly different from those used for adversarial robustness (see e.g.\ \cite{Tjeng2019EvaluatingProgramming}). First, rather than looking for perturbations 
around a fixed a point, here we have both $x'$ and $x''$ as variables.
Furthermore, rather than being local, the MILP problem for $\epsilon$-$\delta$-IF is global, over the whole input space $X$.
As such, local approximations of non-linearities cannot be used, as the bounding needs to be valid simultaneously over the whole input space.
Finally, while in adversarial robustness one can ignore the last sigmoid layer, for IF, because of the two optimisation variables, one cannot simply map from the last pre-activation value to the class probability, so that even for ReLU NNs one needs to employ bounding of non-piecewise activations for the final sigmoid.

By combining the results from this section, we have:
\begin{theorem}\label{th:theorem1} Consider $\epsilon \geq 0$, a similarity $d_{\text{fair}}$ and a NN $f^{w}$.
Let $x'_*$ and $x''_*$ be the optimal points for the optimisation problem in Equation \eqref{eq:milp}. 
Define $\delta_* = | f^w(x'_*) - f^w(x''_*) | $. 
Then $f^w$ is $\epsilon$-$\delta$-individually fair w.r.t.\  $d_{\text{fair}}$ for any $\delta \geq \delta_*$.
\end{theorem}

Theorem \ref{th:theorem1}, whose proof can be found in Appendix~\ref{sec:milp_details}, states that a solution of the MILP problem provides us with a sound estimation of individual fairness of an NN.
Crucially, it can be shown that branch-and-bound techniques for the solution of MILP problems converge in finite time to the optimal solution \citep{del2012convergence}, while furthermore providing us with upper and lower bounds for the optimal value at each iteration step. Therefore, we have:
\begin{corollary}\label{th:corollary}
Let $\delta_k^L$ and $\delta_k^U$ lower and upper bounds computed by a MILP solver at step $k >0$. Then we have that:  $\delta_k^L \leq \delta_* \leq \delta_k^U $. Furthermore, given a precision, $\tau$, there exist a finite $k_*$ such that $\delta_{k_*}^U - \delta_{k_*}^L \leq \tau   $.
\end{corollary}

That is, our method is sound and anytime, as at each iteration step in the MILP solving we can retrieve a lower and an upper bound on $\delta_*$, which can thus be used to provide provable guarantees while converging to $\delta_*$ in finite time.


\paragraph{Complexity Analysis}
The 
encoding of the model constraint can be done in $O(LMn_{\max})$, where $n_{\max}$ is the maximum width of $f^w$, $L$ is the number of layers, and $M$ is the number of grid points used for the PWL bound.
The computational complexity of the fairness constraints depends on the similarity metric employed.
While for  $\ell_\infty$ no processing needs to be done, the computational complexity is $O(n^3)$ for the Mahalanobis distance and again $O(LMn_{\max})$ for the feature embedding metric.
Each iteration of the MILP solver entails the solution of a linear programming problem and is hence $O((M n_{\max}L)^3)$.
Finite time convergence of the MILP solver to $\delta^*$ with precision $\tau$ is exponential in the number of problem variables, in $\tau$ and  $\epsilon$.

%% file: sections/4_training.tex
\subsection{Fairness Training for Neural Networks}\label{sec:training_reduced}

The $\epsilon$-$\delta$-IF MILP formulation introduced in Section \ref{sec:MILP} can be adapted for the solution of Problem \ref{prob:training}. 
The key step is the computation of $x^*_i$ in the second component of the modified loss introduced in Problem \ref{prob:training}, which is used to introduce fairness directly into the loss of the neural network.
This computation can be done by observing that, for every training point $x_i$ drawn from $\mathcal{D}$, the computation of $x^*_i=\arg\max_{x \in X \, s.t.\, d_{x}(x_i,x) \leq \epsilon} |f^w(x_i) - f^w(x)|$ is a particular case of the formulation described in Section \ref{sec:MILP}, where, instead of having two variable input points, only one input point is a problem variable while the other is given and drawn from the training dataset $\mathcal{D}$.
Therefore, $x^*_i$ can be computed by solving the MILP problem, where we fix a set of the problem variables to $x_i$, and can be subsequently used to obtain the value of the modified loss function. Note that these constraints are not cumulative, since they are built for each mini-batch, and discarded after optimization is solved to update the weights.

\begin{algorithm}[h]\footnotesize
\caption{Fair Training with MILP.}\label{alg:FairTraining}
\textbf{Input:} NN architecture: $f^w$, Dataset: $\mathcal{D}$, Learning rate: $\alpha$, Iterations: $n_{\text{epoch}}$, Batch Size: $n_{\text{batch}}$, Similarity metric: $d_{\text{fair}}$,  Maximum similarity: $\epsilon$, Fairness Loss Weighting: $\lambda$.\\
\textbf{Output:} $w_{\text{fair}}$: weight values balancing between accuracy and fairness.\\B
\vspace*{-0.35cm}
\begin{algorithmic}[1]
\STATE $w_{\text{fair}} \gets InitWeights(f^w)$
\FOR{$t = 1,\ldots,n_{\text{epoch}}$}
  \FOR{$b = 1,\ldots,  \lceil |\mathcal{D}| / n_{\text{batch}} \rceil$}
    \STATE $\{X, Y\} \gets \{x_i, y_i\}_{i=0}^{n_{\text{batch}}} \sim \mathcal{D}$ \hfill{ \#Sample Batch}
    \STATE ${Y}_{\text{clean}} \gets f^w(X)$ \hfill{ \#Standard forward pass}
    \STATE $[\bm{\phi}', \bm{\zeta}', \bm{\phi}'', \bm{\zeta}''] \gets  InitMILP(f^w,d_{\text{fair}},\epsilon)$ \hfill{\# Section \ref{sec:MILP}}
    \STATE $X_{\text{MILP}} \gets \emptyset $
    \FOR{$i = 0, \ldots n_{\text{batch}} $}
        \STATE $\bm{\phi}'_i, \bm{\zeta}'_i \gets  FixVarConst(x_i) $ \hfill{ \#Fix constraints}
        \STATE $x^{*}_{i} \gets MILP(x_i, \bm{\phi}'_i, \bm{\zeta}'_i)$ \hfill{\# Solve `local' MILP prob.}
        \STATE $X_{\text{MILP}}  \gets X_{\text{MILP}} \bigcup \{x^*_{i}\}$
    \ENDFOR
    \STATE $Y_{\text{MILP}} \gets f^w(X_{\text{MILP}})$ \hfill{ \#MILP inputs forward pass}
    \STATE 
            $l \gets L_{\text{fair}}({Y}_{\text{clean}}, Y, Y_{\text{MILP}},\lambda)$ \hfill{ \#Fair Loss}
    \STATE $w_{\text{fair}} \gets w_{\text{fair}} - \alpha \nabla_{w} l $ \hfill{ \#Optimizer step (here, SGD)}
  \ENDFOR
\ENDFOR
\STATE return $w_{\text{fair}}$ \hfill{ \#Weights optimized for fairness \& accuracy}
\end{algorithmic}
\end{algorithm}

We summarise our fairness training method in Algorithm~\ref{alg:FairTraining}.
For each batch in each of the $n_{\text{epoch}}$ training epochs, we perform a forward pass of the NN to obtain the output, $Y_{\text{clean}}$ (line 5).
We then formulate the MILP problem as in Section \ref{sec:MILP} (line 6), and initialise an empty set variable to collect the solutions to the various sub-problems (line 7). 
Then, for each training point $x_i$ in the mini-batch, we fix the MILP constraints to the variables associated with $x_i$ (line 9), solve the resulting MILP for  $x^*_i$, and place $x^*_i$ in the set that collects the solutions, i.e.\ $X_{\text{MILP}}$.
Finally, we compute the NN predictions on $X_{\text{MILP}}$ (line 13); the result is used to compute the modified loss function (line 14) and the weights are updated by taking a step of gradient descent.
The resulting set of weights $w_{\text{fair}}$ balances the empirical accuracy and fairness around the training points.

The choice of $\lambda$ affects the relative importance of standard training w.r.t. the fairness constraint: $\lambda = 1$ is equivalent to standard training, while $\lambda = 0$ only optimises for fairness.
In our experiments we keep $\lambda = 1$ for half of the training epochs, and then change it to $\lambda = 0.5$.

%% file: sections/5_experiments.tex
\section{Experiments}\label{sec:ExperimentalResults}

In this section, we empirically validate the effectiveness of our MILP formulation for computing $\epsilon$-$\delta$-IF guarantees as well as for fairness training of NNs. 
We perform our experiments on four UCI datasets \citep{UCIDatasets}: the \textit{Adult} dataset (predicting income), the \textit{Credit} dataset (predicting payment defaults), the \textit{German} dataset (predicting credit risk) and the \textit{Crime} dataset (predicting violent crime). 
In each case, features encoding information regarding gender or race are considered sensitive. 
In the certification experiments we employ a 
 precision $\tau$ for the MILP solvers of $10^{-5}$ and a time cutoff of $180$ seconds. 
We compare our training approach with two different learning methods:  \textit{Fairness-Through-Unawareness} (FTU), in which the sensitive features are simply removed, and SenSR \citep{yurochkin2020training}. 
%
%
Exploration of the cutoff, group fairness, certification of additional NNs, scalability of the methods and additional details on the experimental settings are given in Appendix~\ref{sec:additional_results} and \ref{sec:exp_settings}.\footnote{An implementation of the method and of the experiments can be found at \href{https://github.com/eliasbenussi/nn-cert-individual-fairness}{https://github.com/eliasbenussi/nn-cert-individual-fairness}.}

\paragraph{Fairness Certification} \label{subsec:experiments:certification}
We analyse the suitability of our method in providing non-trivial certificates on $\epsilon$-$\delta$-IF with respect to the similarity threshold  $\epsilon$ (which we vary from $0.01$ to $0.25$), the similarity metric  $d_{\text{fair}}$ , the width of the NN (from $8$ to $64$), and its number of layers (from $1$ to $4$). These reflect the characteristics of NNs and metrics used in the IF literature \citep{yurochkin2020training, ruoss2020learning, Urban2020PerfectlyNetworks}; for experiments on larger architectures, demonstrating the scalability of our approach, see Appendix~\ref{apx:larger-verification}.
For each dataset we train the NNs by employing the FTU approach. 

\begin{figure}
    \centering
    \includegraphics[width=0.41\textwidth]{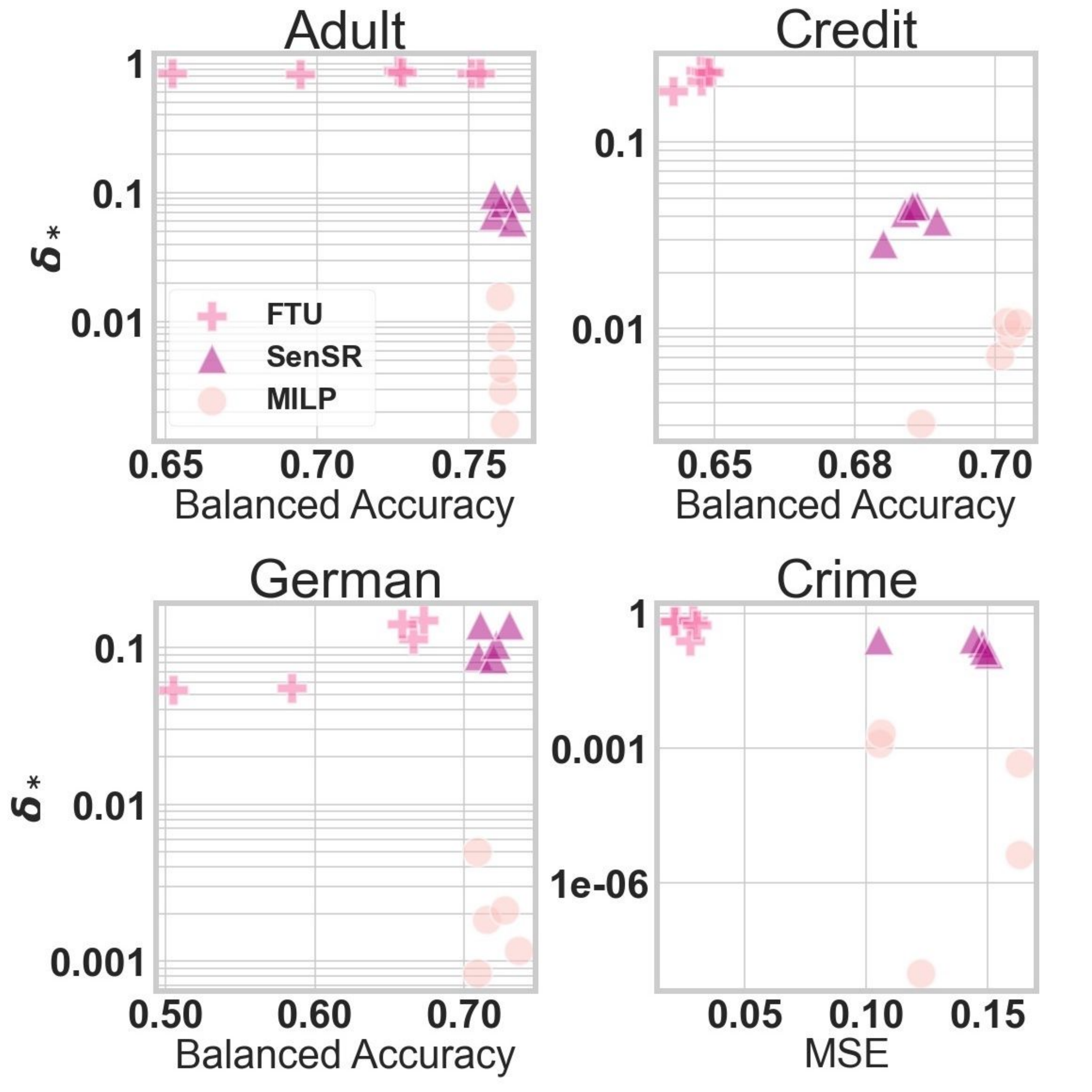} 
    \caption{ Balanced accuracy / individual fairness trade-off for NNs.} 
    \label{fig:training-fig}
\end{figure}

The results for these analyses are plotted in Figure \ref{fig:verification-fig} for the Adult and the Crime datasets (results for Credit and German datasets can be found in Appendix~\ref{apx:credit-german-results}).
Each heat map depicts the variation of $\delta_*$ as a function of $\epsilon$ and the NN architecture. The top row in the figure was computed by considering the Mahalanobis similarity metric; the bottom row was computed for a weighted $\ell_{\infty}$ metric (with coefficients chosen as in \cite{john2020verifying}) and results for the feature embedding metrics are given in Appendix~\ref{sec:feature_embdedding}.  
As one might expect, we observe that,  across all the datasets and architectures, increasing $\epsilon$ correlates with an increase in the values for $\delta_*$, as higher values of $\epsilon$ allow for greater feature changes.
Interestingly, $\delta_*$ tends to decrease (i.e., the NN becomes more fair) as we increase the number of NN layers.
This is the opposite to what is observed for the adversarial robustness, where increased capacity generally implies more fragile models \citep{madry2017towards}.
In fact, as those NNs are trained via FTU, the main sensitive features are not accessible to the NN.
A possible explanation is that, as the number of layers increases, the NN’s dependency on the specific value of each feature diminishes, and the output becomes dependent on their nonlinear combination. The result suggests that over-parametrised NNs could be more adept at solving fair tasks -- at least for IF definitions -- though this would come with a loss of model interpretability, and exploration would be needed to assess under which condition this holds.
%
Finally, we observe  that our analysis confirms how FTU training is generally insufficient in providing fairness on the model behaviour for $\epsilon$-$\delta$-IF. 
For each model, individuals that are dissimilar by $\epsilon \geq 0.25$ can already yield a $\delta_* > 0.5$, meaning they would get assigned to different classes if one was using the standard classification threshold of $0.5$. 
\paragraph{Fairness Training} \label{subsec:experiments:training}

We investigate the behaviour of our fairness training algorithm for improving $\epsilon$-$\delta$-IF of NNs. We compare our method with FTU and SenSR \citep{yurochkin2020training}.
For ease of comparison, in the rest of this section we measure fairness with $d_{\text{fair}}$ equal to the Mahalanobis similarity metric, with $\epsilon = 0.2 $, for which SenSR was developed. 
%
\begin{figure}
    \centering
    \includegraphics[width=0.41\textwidth]{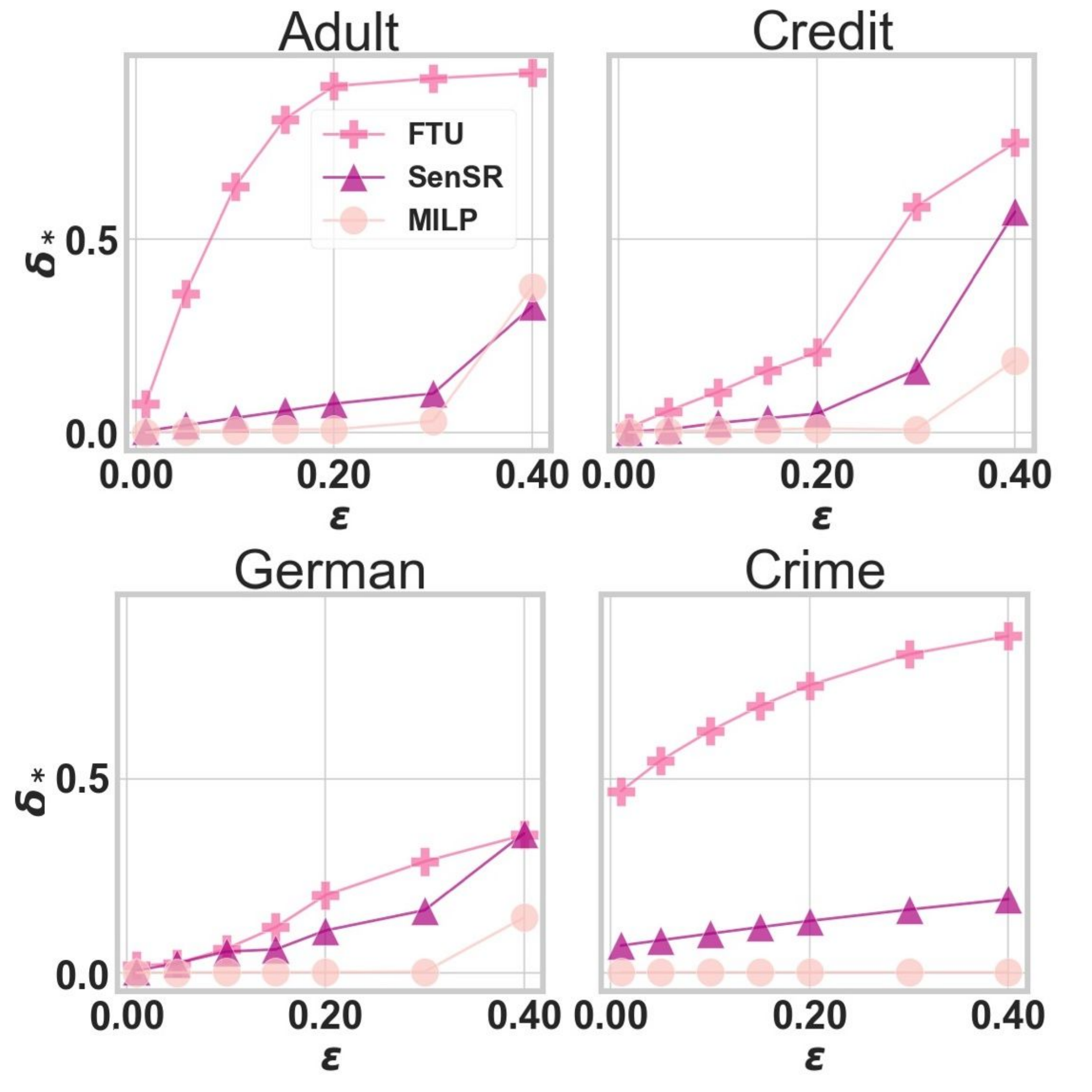}
    \caption{
    Certified $\delta_*$ as a function of the maximum similarity $\epsilon$. 
    }
    \label{fig:training-fig2}
\end{figure}
The results for this analysis are given in Figure \ref{fig:training-fig}, where each point in the scatter plot represents the values obtained for a given NN architecture. We train architectures with up to $2$ hidden layers and $64$ units, in order to be comparable to those trained by \cite{yurochkin2020training}.
As expected, we observe that FTU performs the worst in terms of certified fairness, as simple omission of the sensitive features is unable to obfuscate latent
dependencies between the sensitive and non-sensitive features. 
As previously reported in the literature, SenSR significantly improves on FTU by accounting for features latent dependencies. 
However, on all four datasets, our MILP-based training methodology consistently improves IF by orders of magnitude across all the architectures when compared to SenSR.
In particular, for the architectures with more than one hidden layer, on average, MILP outperforms FTU by a factor of $78598$ and SenSR by $27739$. 
Intuitively, while SenSR and our approach have a similar formulation, the former is based on gradient optimisation so that no guarantees are provided in the worst case for the training loss. 
In contrast, by relying on MILP, our method optimises the worst-case behaviour of the NN at each step, which further encourages training of individually fair models. 
The cost of the markedly improved guarantees is, of course, a  higher computational costs. In fact, the training of the models in Figure~\ref{fig:training-fig} with MILP had an average training time of about $3$ hours.
While the increased cost is significant, we highlight that this is a cost that is only paid once and may be justified in sensitive applications by the necessity of fairness at deployment time. 
We furthermore notice that, while our implementation is sequential, parallel per-batch solution of the MILP problems during training would markedly reduce the computational time and leave for future work the parallelisation and tensorisation of the techniques.   
Interestingly, we find that balanced accuracy also slightly improved with SenSR and MILP training in the tasks considered here, possibly as a result of the bias in the class labels w.r.t.\ sensitive features. 
Finally, in Figure \ref{fig:training-fig2} we further analyse the certified $\delta_*$-profile w.r.t.\ to the input similarity $\epsilon$, varying the value of $\epsilon$ used in for the ceritification of $\epsilon$-$\delta$-IF. In the experiment, both SenSR and MILP are trained with $\epsilon = 0.2$, which means that our method, based on formal IF certificates, is guaranteed to outperform SenSR up until $\epsilon = 0.2$ (as in fact is the case). 
Beyond $0.2$, no such statement can be made, and it is still theoretically possible for SenSR to outperform MILP in particular circumstances. Empirically, however, MILP-based training still largely outperforms SenSR in terms of certified fairness obtained.

%% file: sections/6_conclusion.tex
\section{Conclusion}
We introduced an \textit{anytime} MILP-based method for the certification and training of $\epsilon$-$\delta$-IF in NNs, based on PWL bounding and MILP encoding 
of non-linearities and similarity metrics. 
In an experimental evaluation comprising four datasets, a selection of widely employed NN architectures and three types of similarity metrics, we empirically found that our method is able to provide the first non-trivial certificates for $\epsilon$-$\delta$-IF in NNs and yields NNs which are, consistently, orders of magnitude more fair than those obtained by a competitive IF training technique. 



\paragraph{Acknowledgements}
This project was funded by the ERC European Union’s Horizon 2020 research and innovation programme (FUN2MODEL, grant agreement No.~834115).

%% file: sections/7_appendix.tex
\newpage
\appendix

\begin{center}
\textbf{\Large{Appendix to: \\ Individual Fairness Guarantees  for Neural Networks}}
\end{center}

In Section \ref{sec:milp_details} we empirically investigate the convergence of the PWL bounds w.r.t.\ $M$ in the sigmoid case, and provide detailed proofs for the statements of propositions and theorem from the main paper.
 In Section \ref{sec:metric_learning} we discuss how the learning of the similarity metric $d_{\text{fair}}$ was performed. 
Section \ref{sec:exp_settings} details the experimental settings used in the paper and briefly describes fairness-through-unawareness and SenSR. 
Finally, additional experimental results on group fairness, verification, and feature embedding metrics are given in Section \ref{sec:additional_results}.

\section{Additional Details on MILP}\label{sec:milp_details}

\subsection{Analysis of Number of Grid Points}
Interestingly, by inspecting the error bounds derived in Proposition~\ref{proposition:M_convergence} we notice how the uniform error of the PWL bounds goes to zero with the product between the inverse of $M$ and the increments of the derivative of $\sigma$ parametrised with the inverse of $M$.
In practice, this means that choosing the interval points of the grid adaptively depending on the values of $\sigma'$ yields improved rate of convergence for the bounds. In fact, in Appendix~\ref{sec:milp_details}, by choosing the grid points in inverse proportion to $M$ in practice, for $M=32$, we have almost perfect overlap of the PWL with $\sigma$. 
We visualised this in Figure~\ref{fig:sigmoid_discretisation} in the main paper, where we plot the lower and upper PWL functions used in our MILP construction (the plots illustrate the explicit case of the sigmoid activation function in the interval $[-5,5]$).
The inflection point in the case of the sigmoid is in the axis origin, so it is straightforward to discretise the x-axis into convex and concave parts of the sigmoid. In particular, we achieve this by using a non-uniform discretisation of the x-axis that follows the y-axis of the plot. Empirically, we found that this provides better bounds than a uniform x-axis discretisation in the case in which $M$ (number of grid points used) is small.
The figures visually show how the bounds converge as $M$ increases.
Already for $M = 32$ the maximum approximation error is of the order of $10^{-5}$, and thus this is the value we utilise in the experiments. 

\paragraph{Proof of Proposition 1}
Consider the $j$-th activation function and the $i$-th layer we want to show that everytime $\zeta^{(i)}_j = \sigma^{(i)}\left(\phi^{(i)}_j\right)$ it follows that there exist values for $\eta^{(i)}_{j,l} $ and  $y^{(i)}_{j,l}$ for $l=1,\ldots,M$,  such that $\zeta^{(i)}_j$ $\phi^{(i)}_j$ satisfies the constraints in the proposition statement. This would imply that the feasible region defined by the latter equation is larger than that defined by $\zeta^{(i)}_j = \sigma^{(i)}\left(\phi^{(i)}_j\right)$, and that it hence provide a safe over-approximation of it.

By using Lemma \ref{lemma:pwl_boudning}, we know that 
\begin{align*}
\zeta^{(i)}_j  = \sigma^{(i)}(\phi^{(i)}_j) \geq \eta^{(i)}_{j,l}  \zeta^{\text{PWL},(i),L}_{j,l-1} + \eta^{(i)}_{j,l+1} \zeta^{\text{PWL},(i),L}_{j,l}, \\ \zeta^{(i)}_j  = \sigma^{(i)}(\phi^{(i)}_j)  \leq \eta^{(i)}_{j,l}  \zeta^{\text{PWL},(i),U}_{j,l-1} + \eta^{(i)}_{j,l+1} \zeta^{\text{PWL},(i),U}_{j,l},
\end{align*}
where we notice that $\eta^{(i)}_{j,l+1} = (1-\eta^{(i)}_{j,l}) $.
By employing the Special Ordered Set (SOS) 2 reformulation of piecewise functions \citep{milano2000benefits}, we then obtain: 
\begin{align*}
     &\sum_{l=1}^M y^{(i)}_{j,l} = 1, \; \sum_{l=1}^M \eta^{(i)}_{j,l} = 1, \\ &\phi^{(i)}_j = \sum_{l=1}^M \phi^{(i)L}_{j,l} \eta^{(i)}_{j,l}, \;
     y^{(i)}_{j,l}  \leq \eta^{(i)}_{j,l}  + \eta^{(i)}_{j,l+1}, \\
     &\sum_{l=1}^M  \zeta^{\text{PWL},(i),L}_{j,l}\eta^{(i)}_{j,l} \leq \zeta^{(i)}_j \\
     & \sum_{l=1}^M  \zeta^{\text{PWL},(i),U}_{j,l}\eta^{(i)}_{j,l} \geq \zeta^{(i)}_j 
\end{align*}
which is equivalent to the Proposition statement.

\begin{table*}[ht]\large
  \centering
  \renewcommand{\arraystretch}{1.2}
  \begin{tabular}{|c||c|c|c|c|c|c|}
\hline
                & \textbf{Model} & \textbf{Learning Rate} & \textbf{Regularization} & \textbf{Epochs} & \textbf{Hidden Layers}                                        \\ \hline \hline
                & \textbf{FTU}   & 0.025                  & 0.0125                  & 35              &                                                               \\ 
\textbf{Adult}  & \textbf{SenSR} & 0.001                  & 0.05                    & 400             & {[}8{]}, {[}16{]}, {[}24{]}, {[}64{]}, {[}8,8{]}, {[}16,16{]} \\ 
                & \textbf{MILP}  & 0.001                  & 0.05                    & 400             &                                                               \\ \hline
                & \textbf{FTU}   & 0.002                  & 0.02                    & 50              &                                                               \\ 
\textbf{Credit} & \textbf{SenSR} & 0.0025                 & 0.04                    & 100             & {[}8{]}, {[}16{]}, {[}24{]}, {[}64{]}, {[}8,8{]}, {[}16,16{]} \\ 
                & \textbf{MILP}  & 0.0025                 & 0.04                    & 100             &                                                               \\ \hline 
                & \textbf{FTU}   & 0.001                  & 0.02                    & 35              &                                                               \\ 
\textbf{German} & \textbf{SenSR} & 0.0025                 & 0.04                    & 250             & {[}8{]}, {[}16{]}, {[}24{]}, {[}64{]}, {[}8,8{]}, {[}16,16{]} \\ 
                & \textbf{MILP}  & 0.0025                 & 0.04                    & 250             &                                                               \\ \hline 
                & \textbf{FTU}   & 0.001                  & 0.02                    & 35              &                                                               \\ 
\textbf{Crime}  & \textbf{SenSR} & 0.025                  & 0.025                   & 100             & {[}8{]}, {[}12{]}, {[}16{]}, {[}24{]}, {[}8,8{]}, {[}16,16{]} \\ 
                & \textbf{MILP}  & 0.025                  & 0.025                   & 100             &                                                               \\ \hline
\end{tabular}

\caption{Hyperparameter values used for the models employed in the analysis in Section~\ref{subsec:experiments:certification} and in the comparison between training methods in Section~\ref{subsec:experiments:training}}
\label{tab:hyperparameters}
\end{table*}

\paragraph{Proof of Proposition 2}
For simplicity of notation, we drop the subscripts and superscripts from the proof, and refer to a general activation of a general hidden layer of the NN $f^w$.

Without loss of generality, assume the non-linearity $\sigma(\phi)$ is convex in  $[\phi_{l},\phi_{l+1}]$,  with $\phi_{l+1} - \phi_{l} = \frac{\phi_{U} - \phi_{L}}{M}$  (the concave case follows specularly from the convex by opportunely considering $-\sigma(\phi)$).

Following the construction discussed in Section \ref{sec:model_constraint}, the lower bound in this case is given by the tangent through the midpoint, i.e., $\sigma^L(\phi) =  \sigma(c) + ( \phi - c ) \sigma'(c) $, where $c = (\phi_{l+1} - \phi_{l})/2$, where $c = (\phi_{l+1} - \phi_{l+1})/2$. 
We consider the lower bounding error $e_1 (\phi) = |\sigma^L(\phi) - \sigma(\phi) |$.
%
By definition of convexity and differentiability of $\sigma$ we have:
\begin{align*}
    \sigma(c) \geq \sigma(\phi_{l}) + (c - \phi_{l}) \sigma' (\phi_{l}).
\end{align*}
Hence, for the error we obtain the following chain of inequalities:
\begin{align*}
    &e_1(\phi)  = \sigma(\phi) - \sigma^L(\phi) = \\
    &\sigma(\phi)  - \sigma(c) - ( \phi - c ) \sigma'(c)  \leq\\
    &- (c -  \phi) \sigma' (\phi) - ( \phi - c ) \sigma'(c) =\\
    &( \phi - c ) ( \sigma' (\phi) -\sigma'(c) ).
\end{align*}
which can be reformulated in terms of $M$:
\begin{align}
    &e_1(\phi) \leq \nonumber  \\ \label{eq:lower_error} &\frac{\phi_{U} - \phi_{L}}{2M}   \left( \sigma' (\phi_{l+1}) - \sigma' \left(\phi_{l+1} -  \frac{\phi_{U} - \phi_{L}}{2M}  \right) \right)
\end{align}

For the upper-bound function, we have: $\sigma^U(\phi) =  \sigma(\phi_{l}) + ( \phi - \phi_{l} ) \frac{\sigma(\phi_{l+1}) - \sigma(\phi_{l})}{\phi_{l+1} - \phi_{l}}$. 
Again by convexity we obtain:
\begin{align*}
    \sigma(\phi) \geq \sigma(\phi_{l})  - \sigma'(\phi_{l} )(\phi - \sigma(\phi_{l})). 
\end{align*}
so that for the error we have the following chain of inequalities:
\begin{align*}
     &e_2(\phi) = \sigma^U(\phi) - \sigma(\phi) = \\
     &  \sigma(\phi_{l}) + ( \phi - \phi_{l} ) \frac{\sigma(\phi_{l+1}) - \sigma(\phi_{l})}{\phi_{l+1} - \phi_{l}} - \sigma(\phi) \leq \\
     & ( \frac{\sigma(\phi_{l+1}) - \sigma(\phi_{l})}{\phi_{l+1} - \phi_{l}} + \sigma'(\phi_{l} )) ( \phi - \phi_{l} ).
\end{align*}
Hence, by rewriting it in terms of $M$, we obtain:
\begin{align} \nonumber
     &e_2(\phi) \leq  \\& \label{eq:upper_error} \left( \frac{\sigma(\phi_{l} + \frac{\phi_{U} - \phi_{L}}{M} ) - \sigma(\phi_{l})}{ \frac{\phi_{U} - \phi_{L}}{M} } + \sigma'(\phi_{l} )\right) \frac{\phi_{U} - \phi_{L}}{M}.
\end{align}

Uniform convergence as $M$ tends to infinity follows straightforwardly from the fact that  Equations \eqref{eq:lower_error} and \eqref{eq:upper_error} are independent of any particular value of $\phi$ and that they tend to zero as $M$ goes to infinity.

\paragraph{Proof of Theorem 1}
The theorem statement follows if we show that the feasible region of the MILP of Equation \eqref{eq:milp} over-approximates the feasible region of the individual fairness optimisation problem whose constraints are given in Equations \eqref{eq:obj_function} and \eqref{eq:con_function}. 
In fact, if this holds then any solution $\delta_*$ of the optimisation problem of Equation \eqref{eq:milp} would provide an upper bound to the solution of Problem \ref{prob:certification}, so that for any $\delta \geq \delta_*$ we would have that $f^w$ is $\epsilon$-$\delta$-IF.

\textit{Fairness Constraint:} For the model constraint, this follows directly from the construction of Section \ref{sec:model_constraint}, so that we have that $ - \frac{\epsilon}{\sqrt{\text{diag}(\Lambda)}} \leq U^Tx' - U^Tx'' \leq \frac{\epsilon}{\sqrt{\text{diag}(\Lambda)}}$ implies $d_{\text{fair}}(x',x'') \leq \epsilon$. 

\textit{Model Constraint:} We first rewrite the NN explicitly by using the notation of Equation \eqref{eq:nn} in $x'$ and $x''$, so that we have $\delta =  \zeta'^{(L)} - \zeta''^{(L)} $, and for $i=1,\ldots,L$:
\begin{align*}
       &\zeta'^{(0)} = x', \quad 
    \phi'^{(i)} = W^{(i)} \zeta'^{(i-1)}  + b'^{(i)}, \quad 
    \zeta'^{(i)} = \sigma^{(i)}\left(\phi'^{(i)}\right) \\
    &\zeta''^{(0)} = x'', \; \,
    \phi''^{(i)} = W^{(i)} \zeta''^{(i-1)}  + b''^{(i)}, \; 
    \zeta''^{(i)} = \sigma^{(i)}\left(\phi''^{(i)}\right).
\end{align*}
The first two constraints in each of the two rows above are already linear constraints, and in this form appear in the MILP formulation. 
For the activation constraints, i.e.\ $\zeta^{\dag(i)}_j = \sigma^{(i)}\left(\phi^{\dag(i)}_j\right)$ for $\dag \in \{ ', ''\}$ and $j=1,\ldots,n_i$, we proceed by computing PWL lower and upper bound functions using Lemma \ref{lemma:pwl_boudning} and converting it into MILP form using Proposition \ref{proposition:milp_pwl}.
This yields the final form of the MILP we obtain.

\begin{figure*}[ht]
  \centering
  {\includegraphics[width=1.0\textwidth]{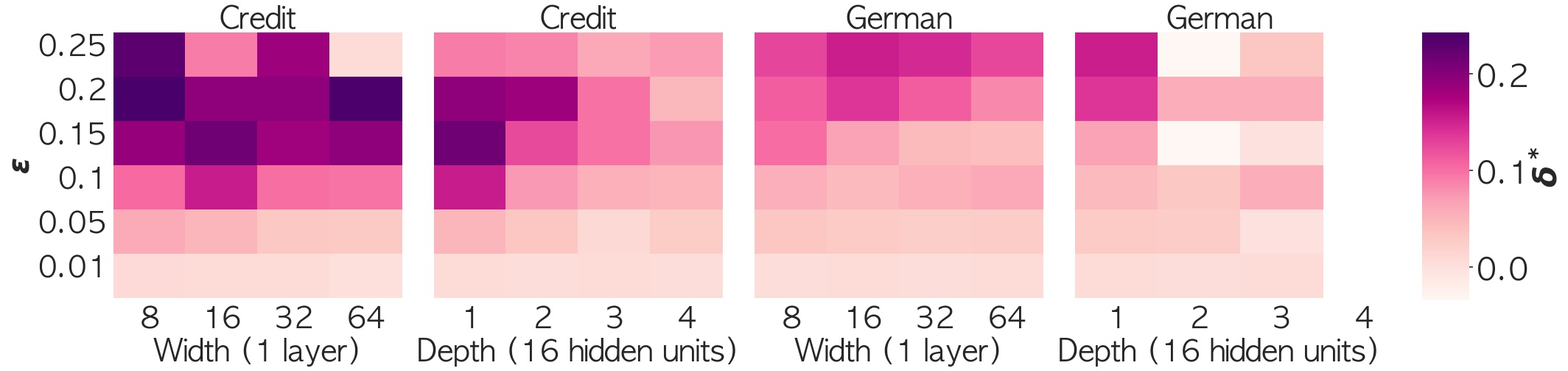}\label{fig:verification-mahalanobis}}
  \hfill
  {\includegraphics[width=1.0\textwidth]{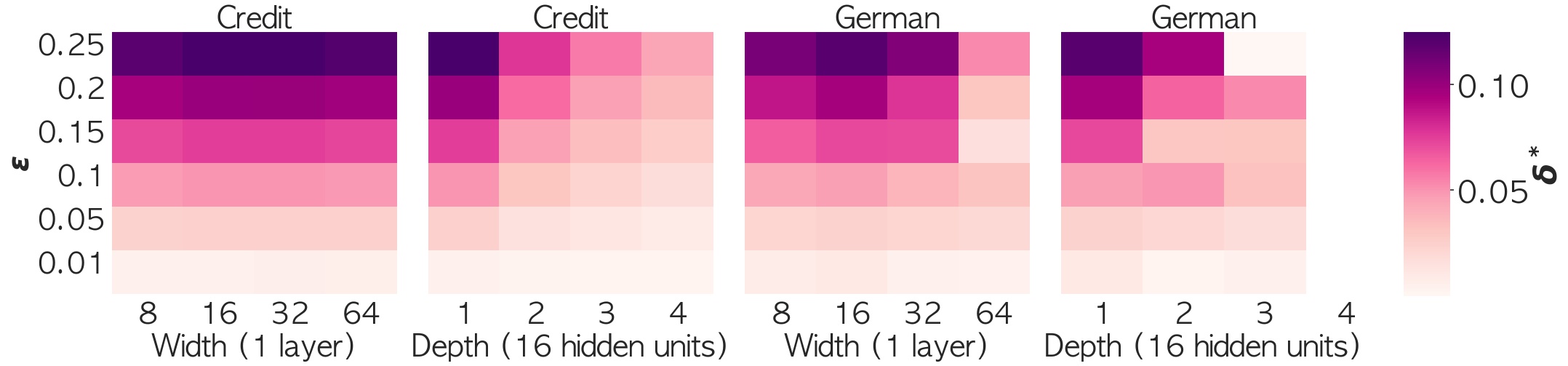}\label{fig:verification-weighted-lp}}
  \caption{Certified bounds on individual fairness ($\delta_*$) for different architecture parameters (widths and depths) and maximum similarity ($\epsilon$) for the Credit (first and second column) and the German (third and fourth column) datasets. \textbf{Top Row}: Mahalanobis similarity metric used for $d_{\text{fair}}$. \textbf{Bottom Row}: Weighted $\ell_{\infty}$ similarity metric used for $d_{\text{fair}}$.
  \label{fig:abl-credit-german}}
\end{figure*}

\section{Metric Learning}\label{sec:metric_learning}
Recently, a line of work aimed at practical methods of learning more expressive fair distance metrics from data has been developed \citep{ilvento19metric, mukherjee20simple, yurochkin2020training}. In this section we expand on the methodology used for metric learning in our experiments.

\subsection{Mahalanobis}\label{sec:mahalanobis_metric}
For the learning of the similarity metric $d_{\text{fair}}$ in the form of a Mahalanobis distance, we rely on the techniques described in \cite{yurochkin2020training} that form the basis of the SenSR approach (to which we compare in our experiments). Briefly, this works as follows.
Consider for simplicity the case of one sensitive feature (e.g., race) with $K$ possible categorical values.
We train a softmax model to predict each value of the sensitive feature by relying on the non-sensitive features.
Let $x_{\text{non-sens}}$ denote the feature vector corresponding to only the non-sensitive features, and similarly $x_{\text{sens}}$ denoting the sensitive features. We then have:
\begin{align}\label{eq:softmax}
    p(x_{\text{sens}} = k ) = \frac{\exp{(a^{T}_{k} x_{\text{non-sens}} + b_{k}})}{\sum_{k=1}^{K} \exp{(a^{T}_{k} x_{\text{non-sens}} + b_{k}})},\; k=1,...,K
\end{align}
where $ p(x_{\text{sens}} = k )$ indicates the confidence given by the softmax model to the sensitive feature having the $k$-th value.
Intuitively, the vector $a^{T}_{k}$, for $k=1,...,K$, then represents a sensitive direction in the non-sensitive features space that correlates $x_{\text{non-sens}}$ to the $k$-th value of $x_{\text{sens}}$.
We then stack the weights of each model, defining the matrix $A = [a_{1},\ldots,a_{K}]$, and compute its matrix span $\text{ran}(A)$, which combines all the sensitive directions in defining a sensitive subspace.
We finally find its orthogonal projector $S = I - P_{\text{ran(A)}}$, which is then used to define the Mahalanobis distance metric as: $d_{\text{fair}}(x',x'') = \sqrt{(x'-x'')^T S (x'-x'')}$.

In the case in which the sensitive feature has a continuous rather than a categorical value, the softmax model of Equation \eqref{eq:softmax} can be replaced by a linear fitting model, and the remainder of the computation follows analogously. 
Finally, we remark that in the case in which many features are selected as sensitive, one can proceed similarly to what has been described just above, by learning a different model for each sensitive feature, and then stacking all the weights obtained together when defining the matrix $A$.

\subsection{Weighted $\ell_p$}
For ease of comparison, we rely on the approach of \cite{john2020verifying}, which in particular focuses on a weighted $\ell_{\infty}$ metric, by setting up the weights to zero for the sensitive features and to a common $\epsilon$ for all the remaining features (we remark that our method is not limited just to $\ell_{\infty}$, but can be used for any general weighted $\ell_p$ metric).
In the experiments described in Section \ref{sec:ExperimentalResults} of the main paper, we consider multiple values for $\epsilon$ varying from $0.01$ to $0.25$. 

\subsection{Feature Embedding}
In addition to the Mahalanobis and  weighted $\ell_p$ distance metric,  we also allow for the metric $d_{\text{fair}}$ to be computed on an embedding. Intuitively, this allows for more flexibility in modelling the intra-relationship between the sensitive and non-sensitive features in each data point and can be used to certify individual fairness in data representations such as those discussed by \citep{ruoss2020learning}.
As a proof of concept, we do this by learning a one-layer neural network embedding of $10$ neurons, and employ the weighted $\ell_{\infty}$ metric. 
Results for this analysis will be given in Section \ref{sec:feature_embdedding}.

\section{Experimental Setting} \label{sec:exp_settings}

In this section we describe the datasets used in this paper and any preprocessing performed prior to training and certification.
We then report the hyperparameter values used to train the different models used in the experiments.
All  experiments were run on a NVIDIA 2080Ti GPU with a 20-core Intel Core Xeon 6230.

\begin{figure}[ht]
  \centering
  \includegraphics[width=0.45\textwidth]{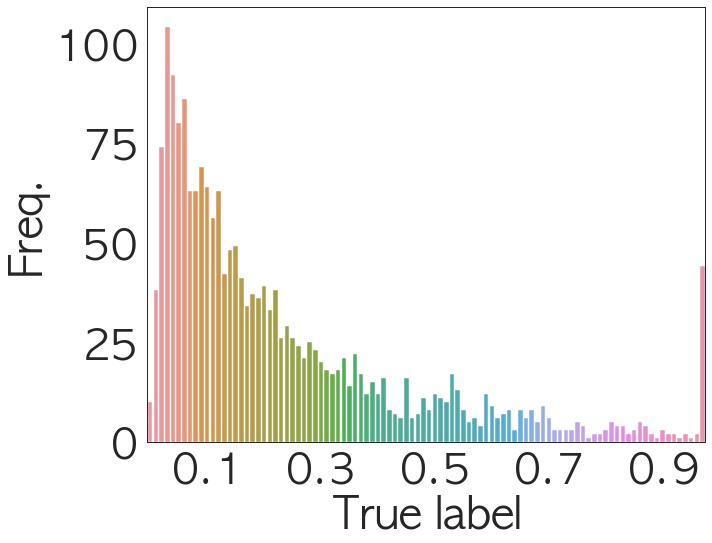}
  \caption{Distribution of the true labels of the Crime dataset. Note how the dataset is very imbalanced towards lower values.}
  \label{fig:crim_dist}
\end{figure}

\subsection{UCI Datasets}\label{sec:datasets}

We consider the following UCI datasets \citep{UCIDatasets}, popular in the fairness literature, with the first three being binary classification tasks and the last one being a regression task.
For all datasets we take an 80/20 train/test split, drop features with missing values, normalise continuous features and one-hot encode categorical features.

\textbf{Adult}: the objective is to classify whether individuals earn more or less than \$50K/year (binary classification). Here we follow similar preprocessing steps as \cite{yurochkin2020training}. After removing \textit{native-country} and \textit{education}, and preprocessing, this dataset contains 40 features, it has 45,222 points, 0.24/0.76 class imbalance, and we consider \textit{sex} and \textit{race} to be categorical sensitive attributes.

\textbf{Credit}: the goal is to predict whether people will default on their payments (binary classification). After preprocessing, the dataset has 144 features, 30,000 data points, a 0.22/0.78 class imbalance, and \textit{x2} (corresponding to sex) is considered a sensitive attribute.

\textbf{German}: the goal is to classify individuals as good/bad credit risks (binary classification). After preprocessing, the dataset has 58 features, 1000 data points, a 0.3/0.7 class imbalance and \textit{status\_sex} is considered a categorical sensitive attribute.

\textbf{Crime}: the goal is to predict the normalised total number of violent crimes per 100K population. After preprocessing, the dataset has 97 features, 1993 data points, and \textit{racepctblack}, \textit{racePctWhite}, \textit{racePctAsian}, \textit{racePctHisp} are considered continuous sensitive attributes. The true label distribution of this dataset is very imbalanced, as shown in Figure~\ref{fig:crim_dist}.

\subsection{Hyperparameters}

The hyperparameters used to train all of the FTU, SenSR and MILP models used in the experiments are reported in Table~\ref{tab:hyperparameters}.
The hidden layer values were selected to match the type of models trained in related literature (e.g. \cite{yurochkin2020training, Urban2020PerfectlyNetworks, ruoss2020learning}). The values of learning rate, regularisation and number of epochs were selected as the result of some hyperparameter tuning, to provide accuracy results matching those found in literature.

\subsection{Training Methods}

Below we describe the alternative fair training methods that are employed for comparison with our proposed training method. We note that for all methods, categorical variables are one-hot encoded, and, since MILP solvers can deal with both continuous and integer variables, no further processing is required.

\paragraph{Fairness through unawarness (FTU)}
The general principle of fairness through unawareness training is that by removing the sensitive features (e.g.\ features containing information about gender or race) the classifier will no longer use such information to make decisions. Despite removal of the sensitive features, it is often found that these have correlations with non-sensitive features, which can lead to classifiers that are still greatly influenced by the sensitive features \citep{pedreshi2008discrimination}.

\paragraph{SenSR}
SenSR is a methodology proposed by \cite{yurochkin2020training} that leverages PGD to generate individually unfair adversarial examples to augment the training procedure. It supports similarity metrics in the form of a Mahalanobis distance, akin to the one we describe in Subsection~\ref{sec:mahalanobis_metric}. We adapt their code to work on both binary classification and regression tasks to compare with our MILP method.
Our MILP method bears many similarity to theirs, hence why we use it for comparison. However, while both our training methods rely on adversarial training to mitigate against unfairness, SenSR does not provide any verification methodology. Furthermore, our MILP training, while being meaningfully more computationally intensive, achieves better local optimisation thus proving upon verification to train models order of magnitude fairer than SenSR.

\section{Additional Experimental Results}\label{sec:additional_results}

In this section we give further empirical evidence supporting the effectiveness of our certification framework as well as our fairness training methodology. We start by giving extended results and discussion on certification for the German and Credit datasets, as well as demonstrate that our method can scale to larger networks than those reported in the main text. We then proceed to do the same for our fairness training algorithm and extend the discussion where appropriate.  

\subsection{Fairness Certification for Credit and German Datasets} \label{apx:credit-german-results}

In Figure~\ref{fig:abl-credit-german} we report similar analyses to that of Figure~\ref{fig:verification-fig} (in main text), illustrating how the values of $\delta_*$ change with respect to changes in $\epsilon$ and number of neurons and hidden layers used for the neural network architecture (using the same parameters as in Section \ref{subsec:experiments:certification}) for Credit and German datasets. 
The top row in the figure was computed considering a Mahalanobis individual similarity metric; the bottom row was computed for a weighted $\ell_{\infty}$ metric. Notice that we obtain similar trends as those discussed for Figure~\ref{fig:verification-fig}.

\begin{figure}
  \centering
  \subfloat[]{\includegraphics[width=0.45\textwidth]{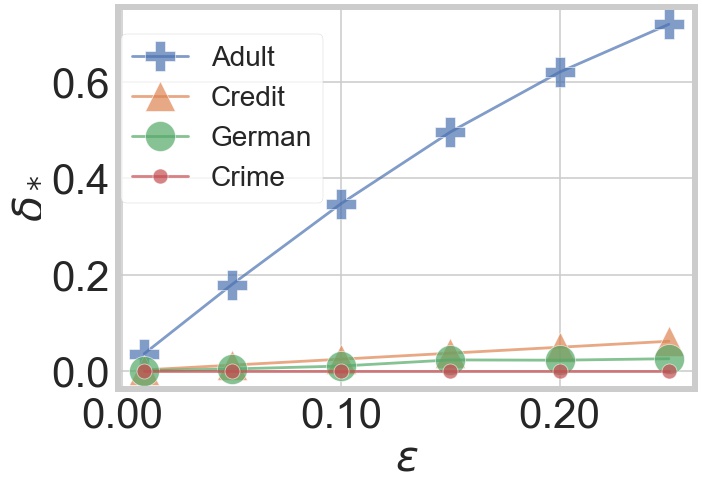}\label{fig:embedding}} \\
  \subfloat[]{\includegraphics[width=0.45\textwidth]{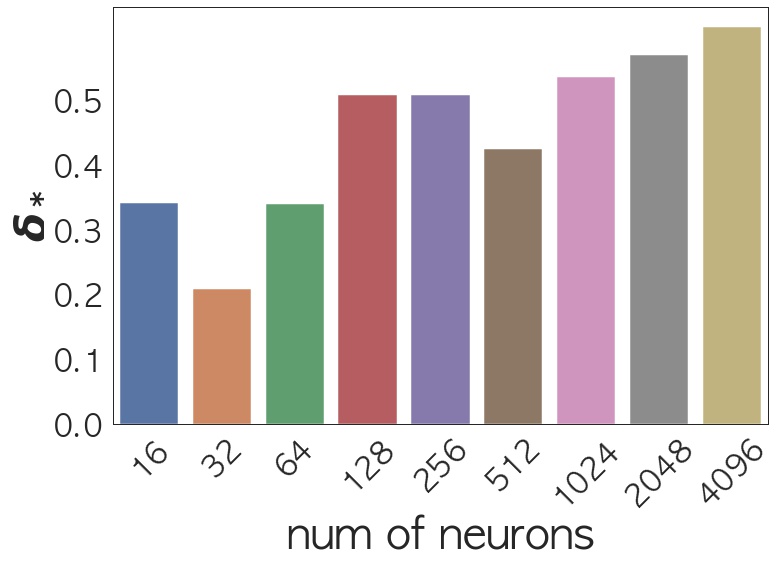}\label{fig:neurons-vs-delta}}
\caption{\ref{fig:embedding} Individual fairness for the various datasets after applying a 10 hidden units linear embedding. All networks are fully connected and have two layers with 16 neurons each. As expected, we observe that  $\delta$ increases with $\epsilon$. Furthermore, the Adult dataset is particularly sensitive to perturbations. \ref{fig:neurons-vs-delta} Certification for increasing architecture size. We notice that our method provides non-vacuous bounds even for hundreds of neurons.} \label{fig:embedding-neurons-vs-delta}
\end{figure}

\subsection{Certification with Feature Embedding Similarity Metric}\label{sec:feature_embdedding}

In Figure~\ref{fig:embedding} we depict the results for individual fairness certification by using the feature embedding approach for the definition of $d_{\text{fair}}$. In particular, we use a neural network with 10 neurons as the embedding. The results are given for a NN with 2 layers, 16 hidden units per layer, using FTU for each dataset. 
The results show that we are able to obtain non-trivial bounds even when an embedding is used.

\subsection{Additional Hyperparameters Analysis} \label{apx:larger-verification}

Up to this point we have studied architectures matching those explored in related literature \citep{yurochkin2020training, ruoss2020learning, Urban2020PerfectlyNetworks}. We now show that our verification method can be applied to far larger architectures to obtain non-trivial results.
In Figure~\ref{fig:neurons-vs-delta} we train with FTU six different neural networks on the Crime dataset, each with 2 hidden layers with the same number of neurons and up to 2048 neurons per layer. For reference, the largest network trained by \cite{yurochkin2020training} has a single hidden layer with 100 neurons, and comes with no formal guarantees.

\begin{figure}
  \centering
  \subfloat[]{\includegraphics[width=0.35\textwidth]{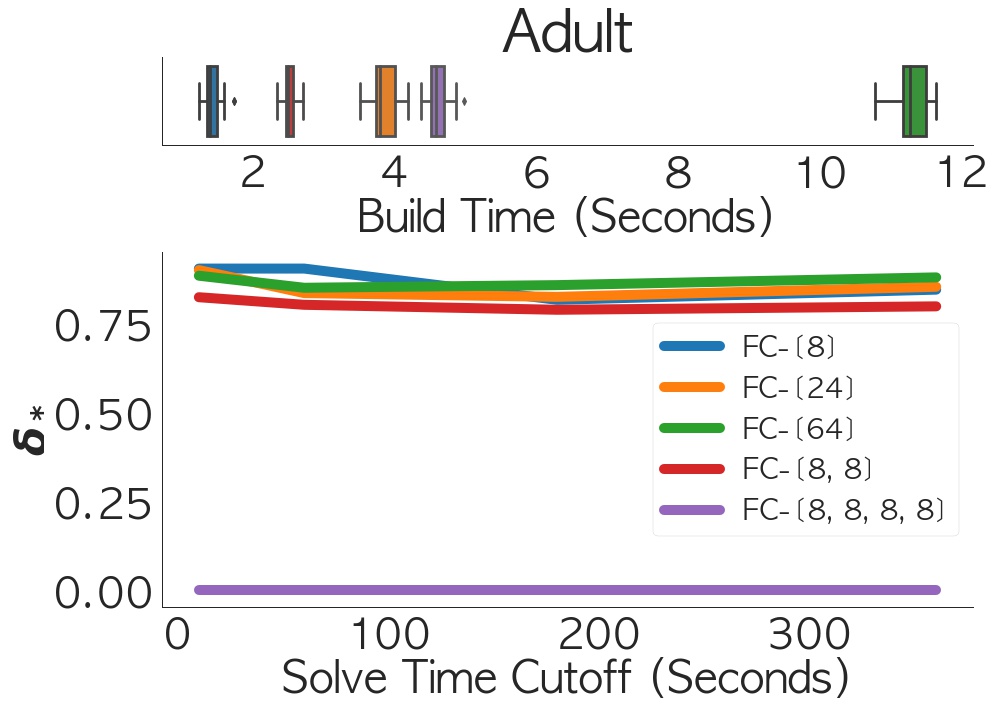}\label{fig:cutoff-adult}}
  \hspace{1cm}
  \subfloat[]{\includegraphics[width=0.35\textwidth]{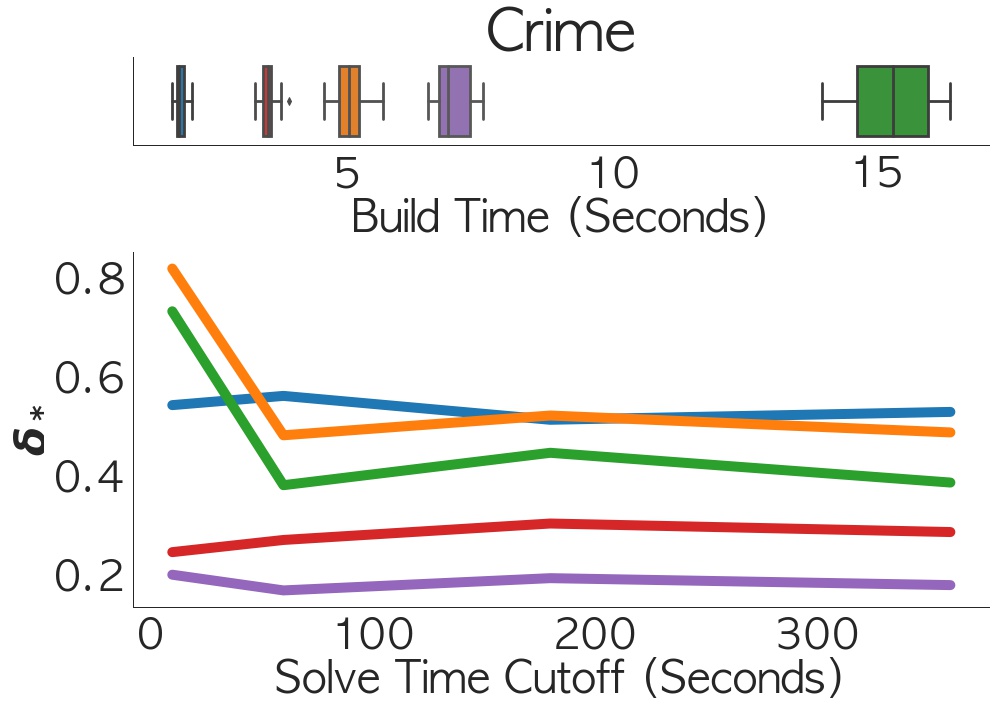}\label{fig:cutoff-crime}}
   \\
  \subfloat[]{\includegraphics[width=0.35\textwidth]{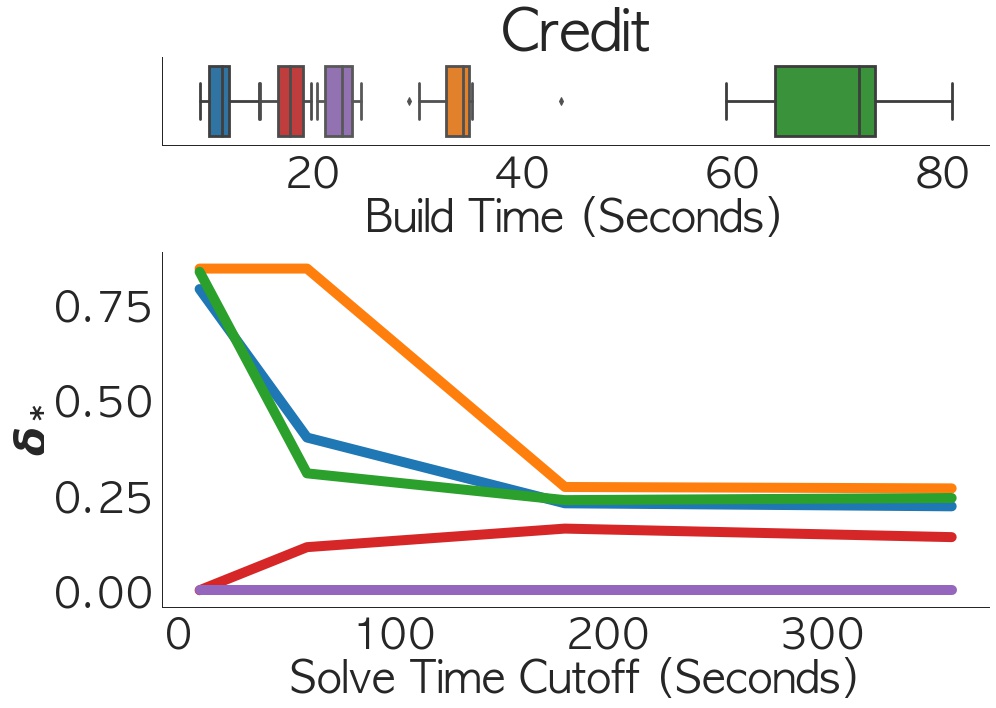}\label{fig:cutoff-credit}}
  \hspace{1cm}
  \subfloat[]{\includegraphics[width=0.35\textwidth]{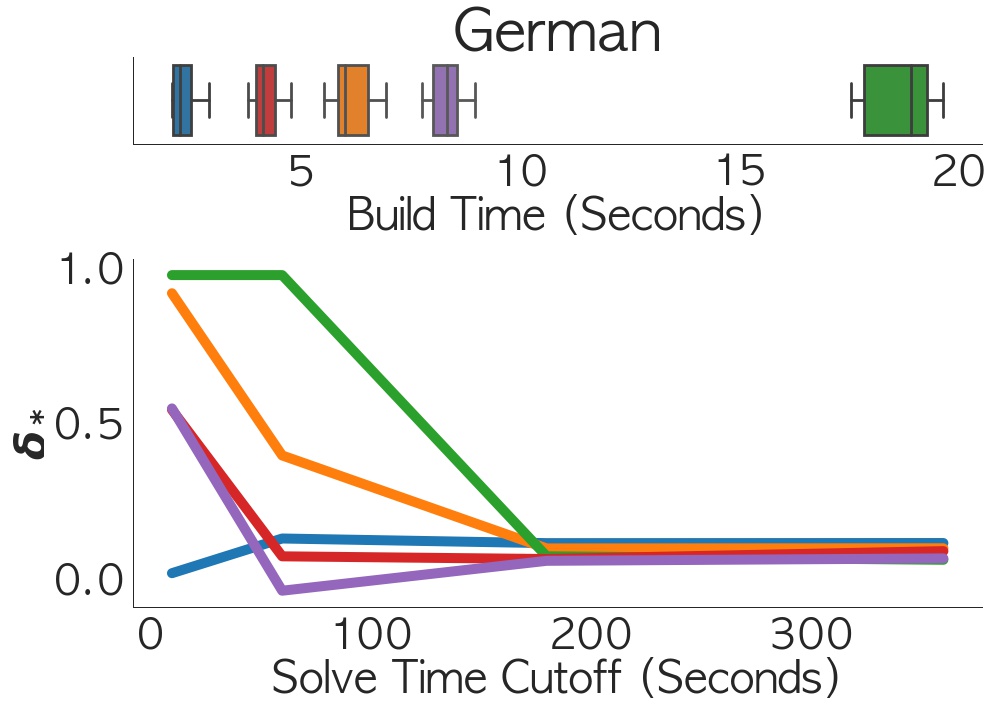}\label{fig:cutoff-german}}
  \caption{In each plot we analyse how the value of delta changes for different cutoff times given to the solver. We analyse this over 5 different architectures.
  The details of the architectures and their corresponding colors are given in the plot (a), the top-most figure. We also report the time taken to build the MILP problem for each architecture, to show how scaling the model size affects each portion of the MILP solution (building and solving constraints).}
  \label{fig:time-cutoff}
\end{figure}

We then report the value of $\delta_*$ certified by the solver for each model (referred to by its total number of neurons, e.g., 16 stands for a NN with 2 hidden layers with 8 neurons each). We notice the value of $\delta_*$ seems to increase. This is most likely due to the fact that for larger networks the MILP problem becomes more difficult to solve, and the time cutoff could has bigger impact on the tightness of the approximation found.
However we note that, even for larger models with thousands of neurons, our verification method yields non-trivial bounds after only 3 minutes of computations.

In Figure~\ref{fig:time-cutoff} we train 5 different fully connected architectures 
for each dataset and verify it for $\epsilon = 0.2$. The details of each architecture and their corresponding colors in the plots are given in the legend of Figure~\ref{fig:time-cutoff}(a), the top most plot of Figure~\ref{fig:time-cutoff}. In particular these architectures vary width and depth. We repeat the verification five times and average the results to mitigate any oscillation due to the solver's inherent randomness. We then observe the effect of varying the cutoff time given to the solver on the value of $\delta_*$. 
We notice that the value of $\delta_*$ appears to converge for all datasets, with the sharpest improvements happening within the first ~200s. Therefore, despite the problem being exponential in the worst case scenario, in practice the over-approximation we implement seems to stabilise relatively quickly, at least empirically in these four datasets.



\begin{figure}[ht]
    \centering
    \includegraphics[width=0.45\textwidth]{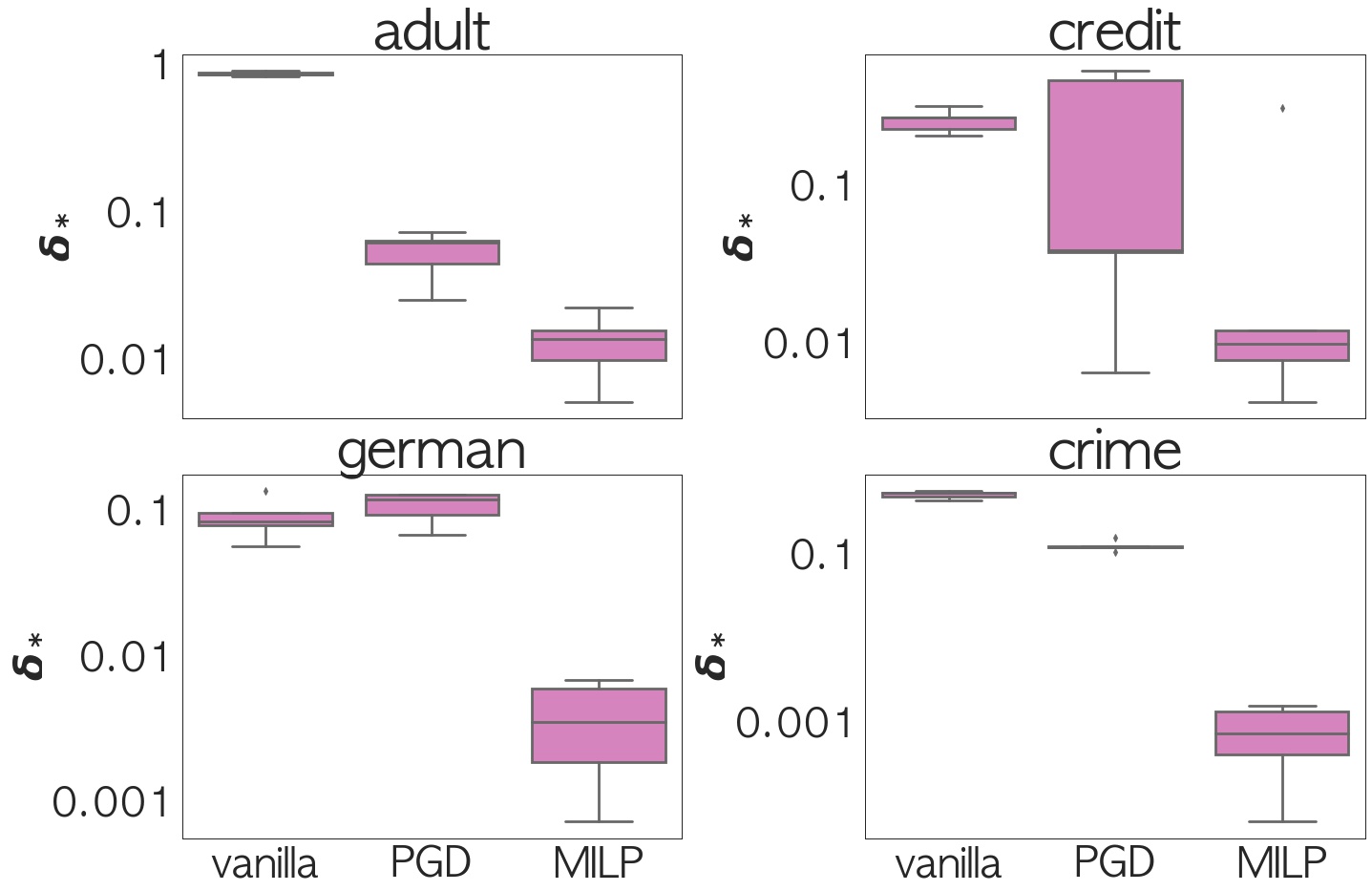}
    \caption{Boxplot showing the quartiles of the distribution of the values of $\delta_*$ for the three models over each dataset training with 5 independent random seeds}
    \label{fig:random-seeds-boxplots}
\end{figure}

\begin{table}[ht]\large
  \centering
  \renewcommand{\arraystretch}{1.2}
  \begin{tabular}{|c||c|c|c|}
    \hline
    & \textbf{FTU} & \textbf{SenSR} & \textbf{MILP} \\
    \hline \hline
    \textbf{Adult} & 0.024 & 0.017 & 0.0056 \\
    \hline
    \textbf{Credit} & 0.038 & 0.22 & 0.12 \\
    \hline
    \textbf{German} & 0.026 & 0.023 & 0.0022 \\
    \hline
    \textbf{Crime} & 0.046 & 0.017 & 0.00055 \\
    \hline
  \end{tabular}
  \caption{Standard deviation values for $\delta_*$ obtained training models with an 8-neurons one hidden layer architecture for the 5 independent random seeds}
  \label{tab:random_seed}
\end{table}

\begin{table}\footnotesize
  \centering
  \renewcommand{\arraystretch}{1.2}
  \begin{tabular}{|c|c||c|c|c|c|}
    \hline
    & \textbf{Model} &\textbf{Adult} & \textbf{Credit} & \textbf{German} & \textbf{Crime} \\
    \hline \hline
                 & \textbf{FTU} & 21s & 28s & 6s & 6s \\
    \textbf{[8]} & \textbf{SenSR} & 17s & 7s & 4s & 3s \\
                 & \textbf{MILP} & 121m & 361m & 12m & 11m \\
    \hline
                  & \textbf{FTU} & - & - & - & 6s \\
    \textbf{[12]} & \textbf{SenSR} & - & - & - & 3s \\
                  & \textbf{MILP} & - & - & - & 14m \\
    \hline
                  & \textbf{FTU} & 22s & 29s & 6s & 6s \\
    \textbf{[16]} & \textbf{SenSR} & 19s & 7s & 4s & 3s \\
                  & \textbf{MILP} & 169m & 354m & 13m & 11m \\
    \hline
                  & \textbf{FTU} & 22s & 28s & 6s & 6s \\
    \textbf{[24]} & \textbf{SenSR} & 21s & 7s & 4s & 3s \\
                  & \textbf{MILP} & 196m & 428m & 15m & 16m \\
    \hline
                  & \textbf{FTU} & 23s & 30s & 6s & - \\
    \textbf{[64]} & \textbf{SenSR} & 22s & 8s & 5s & - \\
                  & \textbf{MILP} & 302m & 628m & 19m & - \\
    \hline
                   & \textbf{FTU} & 23s & 28s & 6s & 6s \\
    \textbf{[8,8]} & \textbf{SenSR} & 20s & 7s & 4s & 3s \\
                   & \textbf{MILP} & 203m & 447m & 14m & 9m \\
    \hline
                     & \textbf{FTU} & 22s & 30s & 6s & 6s \\
    \textbf{[16,16]} & \textbf{SenSR} & 22s & 9s & 5s & 4s \\
                     & \textbf{MILP} & 353m & 710m & 19m & 20m \\
    \hline

  \end{tabular}
  \caption{Training times for different architectures on the various datasets used for Algorithm \ref{alg:FairTraining} using $\epsilon=0.2$}
  \label{tab:training_times}
\end{table}

\subsection{Training}

\begin{figure*}
    \centering
    \includegraphics[width=1.0\textwidth]{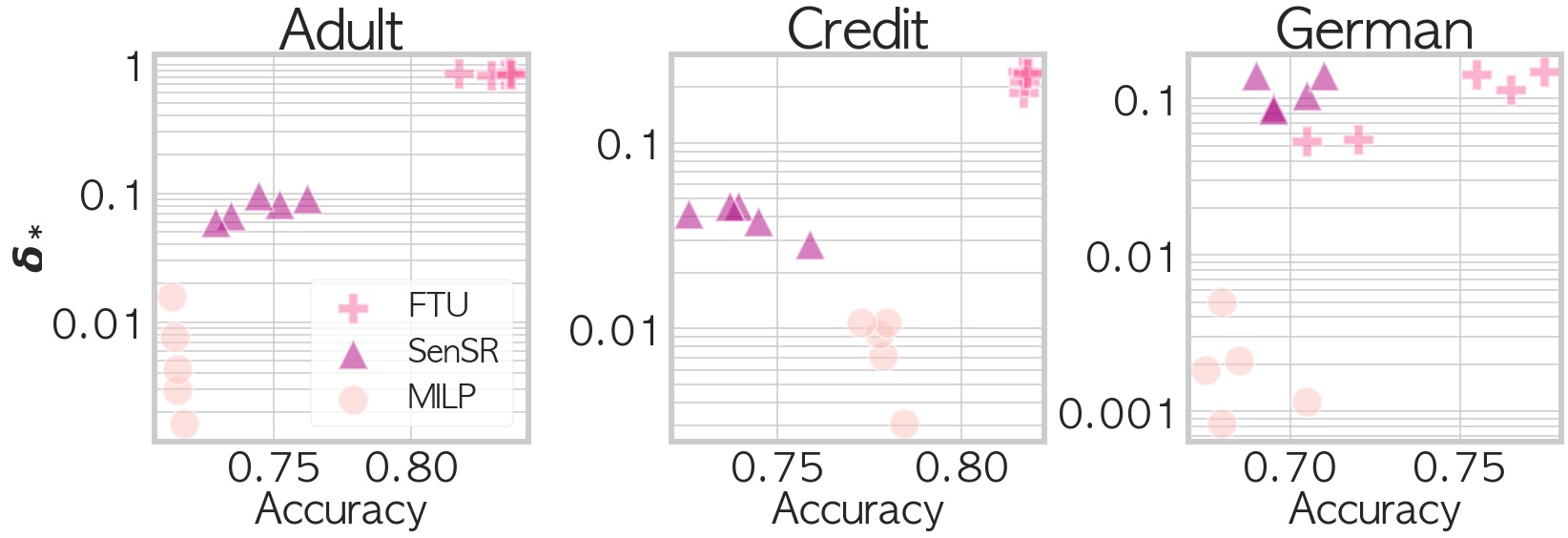}
    \caption{
    Accuracy / individual fairness trade-off for NNs trained with fairness-by-unawareness, SenSR, and MILP for $\epsilon = 0.2$. Each dot in the scatter plot represents a different architecture. }
    \label{fig:accuracy-delta}
\end{figure*}

In Figure~\ref{fig:accuracy-delta} we re-plot the scatter plots from Figure~\ref{fig:training-fig2} (in the main text) for the binary classification tasks using the standard test accuracy on the x-axis (instead of the balanced accuracy).
Notice that, in terms of standard accuracy, FTU clearly outperforms the other two methods.
Intuitively, FTU is only concerned with the maximisation of accuracy, while the other two methods (SenSR and our MILP approach) also optimise for fairness, which, like discussed in Section~\ref{subsec:experiments:training} of the main paper, seems to provide them some regularisation against class imbalance.
We also study the standard deviation in the values of $\delta_*$ training models with different random seeds in Table \ref{tab:random_seed}, and make two observations. Firstly, given those values the difference in values of delta observed for each method remains significant. Secondly, while each of these methods is affected by randomness, our method obtains the smallest values of standard deviation, thus providing more consistent results.
However, as mentioned in the main paper, our MILP based training method has the main drawback of computational overhead. While in the next section we perform further experiments on the scalability of our training method, in Table~\ref{tab:training_times} we report the training times for all the NNs we trained in our experiments, to show that the trade off for orders of magnitude better fairness is significantly longer training times.

\subsection{Training Scalability}


We train a model with the same architecture as the largest network trained by \cite{yurochkin2020training}, with a single hidden layer and 100 neurons, on the Adult dataset (using same values of learning rate, regularization and number of epochs as in Table~\ref{tab:hyperparameters}). Upon verification, our model is guaranteed to have $\delta_* = 0.0007221$, while the same model trained with the methodology from \cite{yurochkin2020training} obtains $\delta_* = 0.042$. This notable improvement in guaranteed fairness bound comes at the significant cost in training time, as our model takes 9h to train. 
We also notice that the guarantees that this network obtains are very similar to the ones obtained with smaller network. 

\subsection{Group Fairness}

In this section, we inspect how our MILP individual fairness training impacts group fairness  as measured by  Equalized Odds Difference (EOD) (calculated using the \href{https://fairlearn.org/}{Fairlearn} library, and inspired by the definition given by \cite{Hardt2016EqualityLearning}). Group fairness definitions largely concern themselves with summary statistics of the model performance on the entire test dataset rather than for an individual, so our training method does not optimize w.r.t. group fairness necessarily, though one would expect more individually-fair models to also improve on group fairness.
We report our results on group fairness in Table~\ref{tab:group_table}.

Interestingly, we observe that our method performs comparably to SenSR and both improve on EOD when compared with FTU.

\begin{table*}\Large
  \centering
  \renewcommand{\arraystretch}{1.2}
  \begin{tabular}{|c||c|c|c|c|c|}
    \hline
    & \textbf{Model} & \textbf{Bal. Accuracy} & \textbf{Accuracy} & \textbf{$\delta_*$} & \textbf{EOD} \\
    \hline \hline
    \multirow{3}{35pt}{\textbf{Adult}} 
     & \textbf{FTU} & 0.718 ± 0.035 & \textbf{0.831 ± 0.007} & 0.848 ± 0.017 & 0.663 ± 0.070 \\
     & \textbf{SenSR} & 0.743 ± 0.012 & 0.694 ± 0.020 & 0.063 ± 0.009 & \textbf{0.550 ± 0.087} \\
     & \textbf{MILP} & \textbf{0.761 ± 0.001} & 0.715 ± 0.002 & \textbf{0.006 ± 0.005} &  0.600 ± 0.000 \\
    \hline
    \multirow{3}{35pt}{\textbf{Credit}} 
     & \textbf{FTU} & 0.648 ± 0.002 & \textbf{0.817 ± 0.001} & 0.230 ± 0.016 & 0.014 ± 0.003 \\
     & \textbf{SenSR} & 0.687 ± 0.005 & 0.743 ± 0.007 & 0.024 ± 0.015 & 0.021 ± 0.011 \\
     & \textbf{MILP} & \textbf{0.699 ± 0.006} & 0.779 ± 0.004 & \textbf{0.008 ± 0.003} & \textbf{0.007 ± 0.004} \\
    \hline
    \multirow{3}{45pt}{\textbf{German }} 
     & \textbf{FTU} & 0.623 ± 0.069 & \textbf{0.751 ± 0.032} & 0.090 ± 0.048 & 0.384 ± 0.181 \\
     & \textbf{SenSR} & \textbf{0.719 ± 0.013} & 0.705 ± 0.009 & 0.107 ± 0.024 & \textbf{0.295 ± 0.006} \\
     & \textbf{MILP} & \textbf{0.719 ± 0.011} & 0.685 ± 0.010 & \textbf{0.002 ± 0.001} & 0.304 ± 0.015 \\
    \hline
  \end{tabular}
  \caption{Comparison of different metrics for 3 different model types across 3 different datasets (all binary classification tasks). The last column is the group fairness metric Equalized Odds Difference (EOD).
  }
  \label{tab:group_table}
\end{table*}